%% file: template.tex
\documentclass{article}

\usepackage{arxiv}

\usepackage[utf8]{inputenc} 
\usepackage[T1]{fontenc}    
\usepackage{hyperref}       
\usepackage{url}            
\usepackage{booktabs}       
\usepackage{amsfonts}       
\usepackage{nicefrac}       
\usepackage{microtype}      
\usepackage{graphicx}
\usepackage[numbers]{natbib}
\usepackage{doi}

\usepackage{multirow}
\usepackage{amsmath}
\usepackage{subcaption}
\usepackage{siunitx}
\DeclareSIUnit \belm {Bm}

\title{From Global to Local: Cluster-Aware Learning for Wi-Fi Fingerprinting Indoor Localisation}

\author{ \href{https://orcid.org/0000-0002-1189-5079}{\includegraphics[scale=0.06]{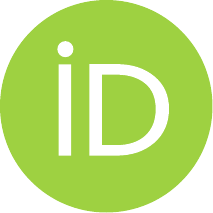}\hspace{1mm}Miguel Matey-Sanz}\\
	Institute of New Imaging Technologies\\
	Universitat Jaume I\\
	Castellón de la Plana, Spain \\
	\texttt{matey@uji.es} \\
	\And
	\href{https://orcid.org/0000-0003-4338-4334}{\includegraphics[scale=0.06]{orcid.pdf}\hspace{1mm}Joaquín Torres-Sospedra} \\
	Departament d'Informàtica\\
	Universitat de València\\
	Burjassot, Spain \\
	\texttt{Joaquin.Torres@uv.es} \\
	\And
	\href{https://orcid.org/0000-0002-8625-441X}{\includegraphics[scale=0.06]{orcid.pdf}\hspace{1mm}Joaquín Huerta} \\
	Institute of New Imaging Technologies\\
	Universitat Jaume I\\
	Castellón de la Plana, Spain \\
	\texttt{huerta@uji.es} \\
	\And
	\href{https://orcid.org/0000-0002-9304-0719}{\includegraphics[scale=0.06]{orcid.pdf}\hspace{1mm}Sergio Trilles} \\
	Institute of New Imaging Technologies\\
	Universitat Jaume I\\
	Castellón de la Plana, Spain \\
	\texttt{trilles@uji.es} \\
}

\renewcommand{\headeright}{Preprint}
\renewcommand{\undertitle}{Preprint}
\renewcommand{\shorttitle}{Cluster-Aware Learning for Wi-Fi Fingerprinting Indoor Localisation}

\hypersetup{
pdftitle={From Global to Local: Cluster-Aware Learning for Wi-Fi Fingerprinting Indoor Localisation},
pdfsubject={cs.ML},
pdfauthor={Miguel Matey-Sanz, Joaquín Torres-Sospedra, Joaquín Huerta, Sergio Trilles},
pdfkeywords={Indoor Positioning, Indoor Localisation, Wi-Fi Fingerprinting, Zone-based localisation, Clustering, Machine Learning},
}

\begin{document}
\maketitle

\begin{abstract}
Wi-Fi fingerprinting remains one of the most practical solutions for indoor positioning, however, its performance is often limited by the size and heterogeneity of fingerprint datasets, strong Received Signal Strength Indicator variability, and the ambiguity introduced in large and multi-floor environments. These factors significantly degrade localisation accuracy, particularly when global models are applied without considering structural constraints. This paper introduces a clustering-based method that structures the fingerprint dataset prior to localisation. Fingerprints are grouped using either spatial or radio features, and clustering can be applied at the building or floor level. In the localisation phase, a clustering estimation procedure based on the strongest access points assigns unseen fingerprints to the most relevant cluster. Localisation is then performed only within the selected clusters, allowing learning models to operate on reduced and more coherent subsets of data. The effectiveness of the method is evaluated on three public datasets and several machine learning models. Results show a consistent reduction in localisation errors, particularly under building-level strategies, but at the cost of reducing the floor detection accuracy. These results demonstrate that explicitly structuring datasets through clustering is an effective and flexible approach for scalable indoor positioning.
\end{abstract}


\section{Introduction}\label{sec:1}
Accurate indoor positioning is a key enabler for a wide range of location-based services, including indoor navigation~\cite{el2021indoor}, emergency response~\cite{tseng2022real}, asset tracking in industrial environments~\cite{silva2021trackinfactory}, and context-aware applications~\cite{ferrato2025cross}. While Global Navigation Satellite Systems (GNSS) provide reliable positioning outdoors, their performance degrades significantly in indoor environments due to signal attenuation and multipath effects. Moreover, people tend to spend more time in indoor environments~\cite{indooratlas}, which might increase the demand for accurate Indoor Positioning Systems (IPS). As a result, alternative technologies --e.g., Wi-Fi, Bluetooth Low Energy (BLE), Ultra-WideBand (UWB), etc.-- and methods --e.g., lateration, fingerprinting, etc.-- have been widely explored~\cite{liu2007survey}, among which Wi-Fi fingerprinting remains one of the most practical and cost-effective solutions due to the pervasive availability of Wi-Fi infrastructure~\cite{khalajmehrabadi2017modern}.

Wi-Fi fingerprinting systems rely on matching the Received Signal Strength Indicator (RSSI) measurements from Access Points (APs) against a pre-collected radio map. Despite their widespread adoption, these systems face several challenges. RSSI measurements are inherently noisy and highly variable due to environmental factors that cause RSSI fluctuations~\cite{bao2024addressing} and degradation~\cite{silva2021quantifying}, device heterogeneity~\cite{lui2011differences}, and human presence~\cite{wattananavin2020comparative}. Moreover, the high dimensionality of fingerprint data, together with the increasing scale and dynamic nature of modern indoor environments, particularly multi-floor buildings, lead to increased computational complexity and ambiguity during localisation~\cite{chia2025challenge}. These factors often limit the accuracy and scalability of global fingerprinting models.

To address some of these challenges, recent research has explored the use of clustering and hierarchical modelling techniques to structure fingerprint datasets prior to localisation. Beyond reducing the search space, these approaches partition the radio map into smaller and more coherent regions based on spatial proximity or radio similarity, allowing learning models to operate on localised subsets of data. This structured organisation implicitly forms communities of fingerprints that share similar contextual characteristics, facilitating more robust, adaptable, and scalable indoor localisation in dynamic environments~\cite{tian2013fingerprint, liu2016coordinate, sadhukhan2021efficient, quezada2025modular}. By enabling localisation models to be trained and applied within these communities, such approaches provide a natural basis for maintaining and updating positioning systems as radio environments evolve over time. Although clustering is performed on fingerprints, the resulting communities exhibit dynamic behaviour due to the continuous movement of devices collecting measurements across the environment.

In this work, we propose a clustering-based method for Wi-Fi fingerprinting that aims to improve indoor positioning accuracy while maintaining flexibility and scalability. The method supports clustering based on either spatial coordinates or RSSI features and can be applied at both building and floor levels. A dedicated cluster estimation mechanism assigns incoming fingerprints to clusters, after which Machine Learning (ML) models are used to estimate position and floor. The proposed method is evaluated using three Wi-Fi fingerprinting datasets and several ML models, showing significant improvements among them compared with a baseline method. The main contributions of the paper are as follows:

\begin{itemize}
    \item We introduce a clustering-based method for Wi-Fi fingerprinting that supports both spatial- and radio-domain clustering, as well as building- and floor-level strategies.
    \item We propose a cluster estimation mechanism, based on the strongest APs, to assign unseen fingerprints to clusters during the localisation phase.
    \item We conduct an extensive experimental evaluation on three public and heterogeneous datasets using multiple ML algorithms, analysing the impact of different clustering strategies on both positioning accuracy and floor detection performance.
\end{itemize}

The remainder of this paper is organised as follows. Section~\ref{sec:2} reviews the background and related work in Wi-Fi fingerprinting and clustering-based localisation. Section~\ref{sec:3} describes the proposed clustering-based positioning method. Section~\ref{sec:4} presents the experimental setup, datasets, and evaluation metrics. Section~\ref{sec:5} and \ref{sec:6} present and discuss the results of the proposed method. Finally, Section~\ref{sec:7} concludes the paper and outlines directions for future work.

\section{State of the art}\label{sec:2}
This section reviews the state of the art in Wi-Fi fingerprint-based indoor localisation. Section~\ref{sec:2_1} introduces the main background concepts, while Section~\ref{sec:2_2} surveys related work focusing on clustering and local modelling approaches.

\subsection{Background}\label{sec:2_1}
IPS have become essential for location-based services in environments where GNSS is unavailable or unreliable, such as inside buildings. Among the various technologies used, Wi-Fi fingerprinting is one of the most widely adopted due to the ubiquitous availability of Wi-Fi infrastructure and its low implementation cost compared to other signals such as UWB or BLE \cite{brena2017evolution, mendoza2019meta}. In Wi-Fi fingerprinting, the RSSI measurements from multiple APs are collected during a training (or offline) phase to construct a radio map, which is then used in a localisation (or online) phase to match new measurements and estimate the user's position using pattern recognition or ML techniques \cite{xia2017indoor}.

Clustering, which is the process of grouping data points based on some specified similarity criteria (e.g., distance measurements, correlations, probabilistic measures, etc.), has also been studied extensively within pattern recognition and ML as a means to reduce complexity and capture structure in datasets, so specialised estimation models can be trained for each similar group of the original dataset~\cite{xu2015comprehensive}. A diverse range of clustering algorithms exists, such as K-Means, which partitions the data into a pre-specified number of groups by minimising within-cluster variance; DBSCAN, a density-based clustering algorithm that groups samples according to local point density~\cite{ester1996density}; or Affinity Propagation, which exchanges similarity messages between data points and determines clusters based on pairwise similarities~\cite{frey2007clustering}. These clustering algorithms can be applied to Wi-Fi fingerprint data to reveal spatial structures when operating on location features, or radio structures when operating on RSSI measurements, which can be exploited to support indoor localisation tasks.

\subsection{Related works}\label{sec:2_2}
Research in Wi-Fi fingerprint-based positioning has explored numerous strategies to enhance accuracy and computational efficiency, including the use of clustering to support more localised modelling approaches. For instance, \citet{torres2020comprehensive} investigated several clustering techniques within $k$-Nearest Neighbours (KNN)-based indoor positioning, showing that K-Means, Affinity Propagation, and other rules based on strongest APs reduce the computational demand while maintaining positioning accuracy.

Clustering approaches based on radio signal similarity have largely been explored in the literature. For example, Affinity Propagation-based clustering has been applied to reduce the search space of reference points and the associated computational cost in RSSI-based systems~\cite{tian2013fingerprint}, and has also been applied along hybrid distances and adjusting clusters, demonstrating improved accuracy over classical strategies~\cite{bi2021improved}. Other studies have investigated alternative clustering techniques, such as in \cite{sadhukhan2021efficient}, where a clustering strategy with outlier mitigation was proposed to reduce computational cost while maintaining positioning performance. A different approach was investigated by \citet{quezada2025modular}, in which clusters were formed based on the strongest APs rather than on full RSSI fingerprints. Their approach showed an improved positioning accuracy in several public datasets, but a decrease in floor classification performance. \citet{hernandez2016hierarchical} also employed the detected APs instead of the fingerprints' RSSI, creating a multi-level hierarchical partitioning of the space based on the visibility of APs and the K-Means algorithm. By grouping fingerprints with similar sets of APs belong to the same cluster, the approach achieved up to a \SI{10}{\percent} improvement in classification accuracy.

Spatial-based clustering methods have also been studied. For instance, \citet{liu2016coordinate} applied the smallest enclosing circle algorithm \cite{chen2012index} over the coordinates of the fingerprints, showing an improvement in positioning accuracy compared to K-Means clustering based on RSSI. \citet{pham2023scalable} also proposed to apply K-Means clustering with the coordinates of the fingerprints to train local estimators which, when combined with a global estimator, achieved a reduced positioning error compared with other methods. Other studies proposed the use of a grid system over fingerprint's coordinates, ranging from one-dimensional partitions~\cite{xiao2025zone} to multi-level hierarchical grids with different levels of granularity, where coarser grids are progressively subdivided into finer ones. These approaches have been shown to improve location accuracy compared with baseline schemes~\cite{abraha2022mriloc}. A similar multi-level hierarchical partitioning strategy was proposed by \citet{kargar2024edge}, where position estimation is performed as a classification task at the finest grid level instead of as coordinate regression. Instead of using clustering techniques, \citet{oh2021c} generated local-building models from a multi-building dataset. This allowed them to include an extra penalty during the training process for exceeding building boundaries, thereby improving the resulting positioning error.

Overall, several clustering strategies play an important role in contemporary indoor positioning research for capturing intrinsic structure in fingerprint data, which can enhance localisation accuracy and scalability across diverse environments.

\section{Methodology}\label{sec:3}
This section describes the proposed method for a clustering-based IPS, composed of two phases: 1) the training phase, during which training data were used to build clusters and train ML models for each cluster, and 2) the localisation phase, in which unseen (testing) data from the same datasets were used to estimate locations and evaluate the performance of the proposed method in terms of positioning error. 

\begin{figure}[h]
    \centering
    \includegraphics[width=\linewidth, trim={.8cm .4cm .8cm .4cm, clip}]{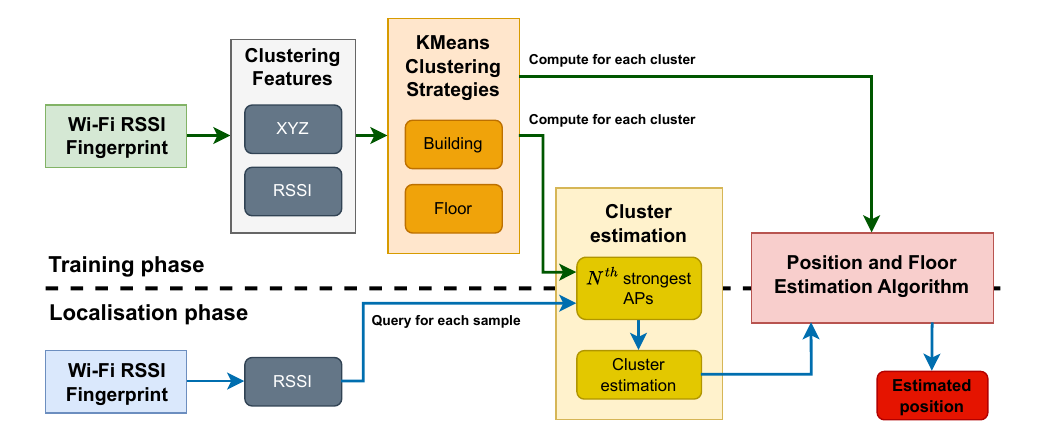}
    \caption{Overall architecture of the proposed clustering-based indoor positioning method, illustrating the training and localisation phases.}
    \label{fig:methodology}
\end{figure}

Figure~\ref{fig:methodology} provides a high-level overview of the proposed clustering-based indoor positioning method, which is organised into a training phase and a localisation phase. In the training phase, Wi-Fi fingerprints are first represented using either spatial or radio features (Section~\ref{subsec:feature_selection}) and then partitioned into clusters using K-Means under different clustering strategies (Section~\ref{subsec:clustering_strategy}). Based on the resulting clusters, a cluster estimation model is built using the strongest APs observed in each cluster (Section~\ref{subsec:cluster_estimation}). During the localisation phase, an unseen fingerprint is assigned to its most relevant cluster, and position and floor estimation are performed using models trained on the selected cluster (Section~\ref{subsec:position_estimation}). Each of these functional blocks is further expanded and detailed in Section~\ref{sec:4}, which describes the experimental setup and the specific configurations associated with each component.

\subsection{Spatial- and radio-based  clustering features}\label{subsec:feature_selection}
\begin{figure}
    \centering
    \begin{subfigure}{0.325\textwidth}
        \includegraphics[width=\linewidth, trim={2.4cm 2.4cm 2.4cm 2cm}, clip]{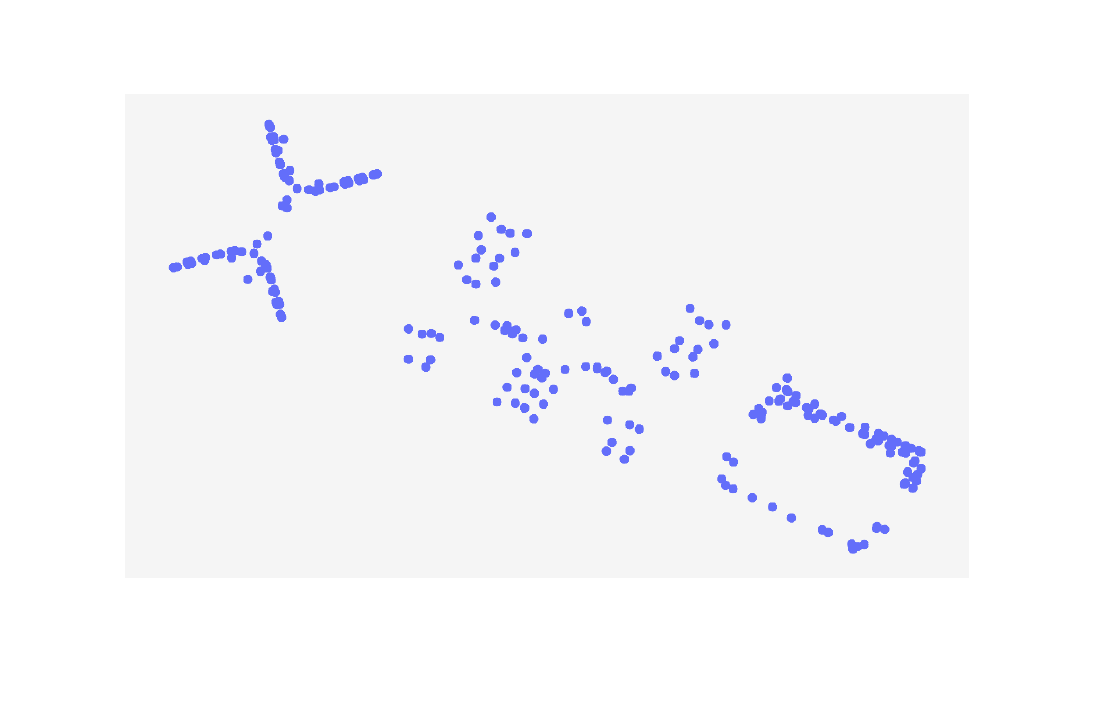}
        \caption{Original fingerprint distribution.}
    \end{subfigure}
    \begin{subfigure}{0.325\textwidth}
        \includegraphics[width=\linewidth, trim={2.4cm 2.4cm 2.4cm 2cm}, clip]{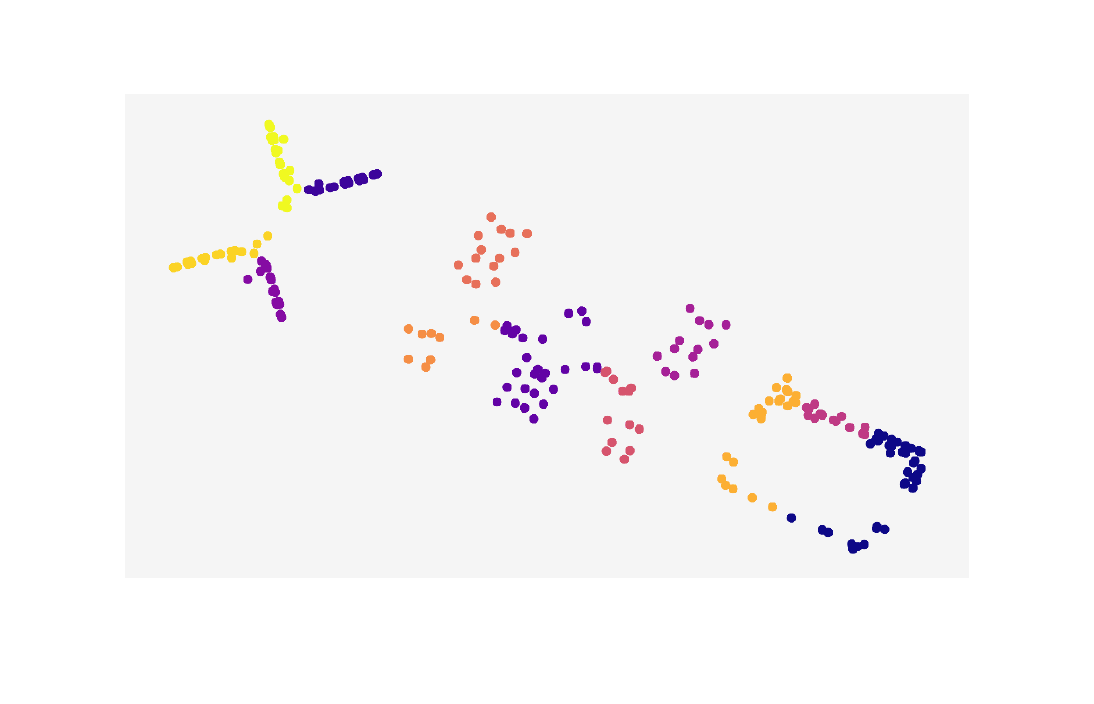}
        \caption{Clusters using spatial (XYZ) features.}
        \label{subfig:xyz_clusters}
    \end{subfigure}
    \begin{subfigure}{0.325\textwidth}
        \includegraphics[width=\linewidth, trim={2.4cm 2.4cm 2.4cm 2cm}, clip]{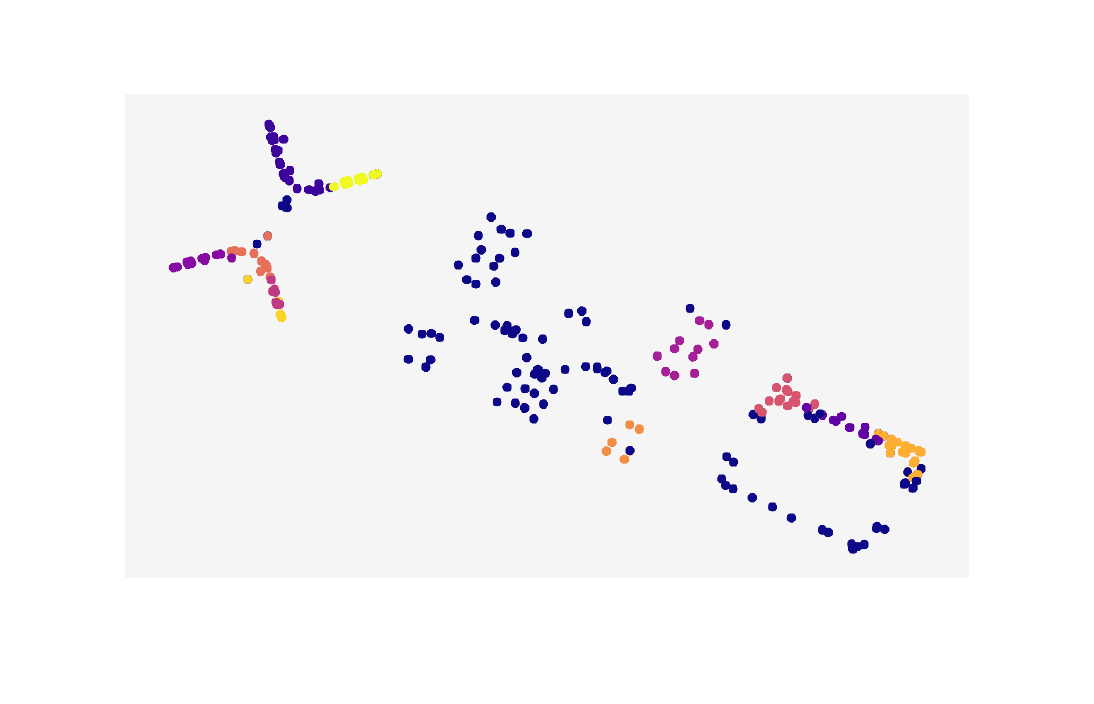}
        \caption{Clusters using radio (RSSI) features.}
        \label{subfig:rssi_clusters}
    \end{subfigure}
    \caption{Comparison of both approaches for creating the clusters in the 2\textsuperscript{nd} floor of UJIIndoorLoc.}
    \label{fig:comparison_cluster_features}
\end{figure}

To enable clustering within the proposed method, two alternative feature representations were considered: spatial-based (namely XYZ) and radio-based (namely RSSI) features. These representations define how Wi-Fi fingerprints are grouped prior to model training, and are illustrated in Figure~\ref{fig:comparison_cluster_features}.

In the first approach, spatial (XYZ) features (see Figure~\ref{subfig:xyz_clusters}), clustering was performed using the ground-truth spatial coordinates of each fingerprint. The clustering algorithm operates in this geometric coordinate space to create groups of fingerprints that are physically close to one another. Therefore, the clusters reflect the spatial structure of the environment, assuming that spatially contiguous regions may share similar propagation conditions and enable local models to perform better than a single global model.

In the second approach, radio (RSSI) features (see Figure~\ref{subfig:rssi_clusters}) were used, and clusters were formed directly from the RSSI measurements. For each $i^{th}$ fingerprint in the dataset, the associated RSSI vector $r_i = [RSSI_{i,1}, \ldots, RSSI_{i,M}]$, where $M$ is the number of APs, is used as the input to the clustering algorithm. Before clustering, RSSI vectors were preprocessed by applying a \textit{powed} representation, which aims to convert the logarithmic scale of RSSI values (\unit{\deci\belm}) to a linear scale in order to work properly with common distance metrics used in several algorithms~\cite{torres2015comprehensive}. This representation is defined by:

\begin{equation}
    Powed_{i,n} = \frac{(RSSI_{i,n} - min)^e}{(-min)^e}\,,
\end{equation}

where $RSSI_{i,n}$ is the RSSI value in the $i^{th}$ fingerprint from the $n^{th}$ AP and $min$ is the minimum global RSSI value of the training dataset 
minus 1, and $e$ is an exponent that controls the non-linear scaling applied to the RSSI values. This approach aims to group fingerprints that exhibit similar radio signal characteristics, regardless of their actual spatial locations. This configuration is intended to capture radio similarity, under the hypothesis that fingerprints with similar RSSI distributions lead to reduced ambiguity during subsequent positioning.

Overall, these two feature representations define complementary clustering approaches within the proposed method. Spatial-based (XYZ) clustering exploits spatial proximity by partitioning the environment into geographically coherent regions, whereas radio-based (RSSI) clustering leverages radio similarity to group fingerprints with comparable signal characteristics. By formulating clusters in either the spatial or radio domain, the method allows restricting model training and inference to subsets of fingerprints that share similar geometric or signal characteristics.

\subsection{K-Means clustering strategies}\label{subsec:clustering_strategy}
The proposed method requires a clustering algorithm to partition the fingerprint dataset based on the selected feature representation. In this work, the K-Means algorithm was adopted due to its simplicity, scalability and suitability for high-dimensional data~\cite{lloyd1982least}. K-Means is an iterative, centroid-based algorithm that groups the dataset into $K$ clusters by minimising the within-cluster sum of squared distances~\cite{ikotun2023k}. The algorithm consists of the following steps:

\begin{enumerate}
    \item Randomly select $K$ initial centroids (i.e., cluster representatives).
    \item Compute the distance from each sample to the centroids and assign the sample to the nearest centroid.
    \item Recompute each cluster centroid as the mean of the samples assigned to the cluster.
    \item Repeat steps 2 and 3 until convergence, i.e., changes in centroid updates are below a certain threshold.
\end{enumerate}

Then, to evaluate how spatial partitioning affects the cluster formation and subsequent positioning performance, each approach described in Section~\ref{subsec:feature_selection} was implemented using K-Means under two different strategies, as described below:

\begin{enumerate}
    \item \textbf{Building-based clustering}: the entire building/dataset was used to train a K-Means model. This strategy allows clusters to form freely across floors, potentially capturing global structure patterns in building geometry (XYZ-features) or radio similarity (RSSI-features).
    \item \textbf{Floor-based clustering}: clustering was performed independently for each floor, producing floor-specific clusters. This strategy constrains clusters to be floor-homogeneous, which can be useful in multi-floor environments where RSSI distributions differ significantly between floors.
\end{enumerate}

While K-Means has several advantages, such as its simplicity, computational efficiency, and scalability, it requires the number of clusters to be specified a priori. The choice of this parameter affects the granularity of the resulting partition and is therefore treated as a configurable component of the proposed method. Although other clustering techniques (e.g., DBSCAN, Affinity Propagation, etc.) could be integrated into the method, their evaluation is outside the scope of this work.

\subsection{Cluster creation and estimation}
\label{subsec:cluster_estimation}
After the clusters are formed in the training phase, a cluster-selection mechanism is required to determine to which cluster an incoming sample should be associated with. This selection is based on $N$ strongest APs observed in the fingerprint.
During the training phase, once clustering is completed and each fingerprint has been assigned to a cluster, the RSSI measurements within each cluster are analysed to construct a set of representative AP combinations. These combinations summarise which groups of strong APs most frequently characterise each cluster. Specifically, for each cluster $C_k$:

\begin{enumerate}
    \item For every fingerprint $f'$ in cluster $C_k$, the $N$ strongest APs are identified: $A_{k,s} = [AP_{1}, AP_{2}, \ldots, AP_{N}]$
    \item Within each cluster, identical AP-combination vectors $A_{k,s}$ may appear multiple times. For cluster $C_k$, the frequency table is defined as:
    \begin{equation}
    \label{eq:cluster_representatives}
        CRep_k = \begin{bmatrix}
        A_{k,1} & freq(A_{k,1}) & C_k \\
        \vdots & \vdots & \vdots \\ 
        A_{k,s} & freq(A_{k,s}) & C_k
        \end{bmatrix}\,,
    \end{equation}
    where $freq(A)$ is the number of fingerprints in $C_k$ whose top-$N$ AP set equals $A$ and $s$ is is the number of distinct top-$N$ AP combinations.    
    \item Finally, all cluster-specific tables are concatenated into a global representation:
    \begin{equation}
    \label{eq:all_representatives}
    ARep = \begin{bmatrix}
        A_{1,1} & freq(A_{1,1}) & C_1 \\
        A_{1,2} & freq(A_{1,2}) & C_1 \\ 
        \vdots & \vdots & \vdots \\ 
        A_{K,S} & freq(A_{K,S}) & C_K
        \end{bmatrix}\,,
    \end{equation}
    where $S$ is the number of distinct top-$N$ AP combinations observed across all $K$ clusters. Therefore, $ARep$ stores every unique AP combination, its frequency, and the cluster where it appears.
\end{enumerate}

During the localisation phase, given a new fingerprint, its $N$ strongest APs are extracted ($A_{test}$). Then, the cluster assignment is derived by comparing this set with the $A_{k,s}$ sets from $ARep$ using the following procedure:

\begin{enumerate}
    \item All non-empty subsets of $A_{test}$ are considered: 
    \begin{equation}
        S_{test} = {all~subsets~of~A_{test}~with~size~1\ldots N}\,, 
    \end{equation}
    resulting in $2^N-1$ subsets.
    \item For each subset $S\in S_{test}$, $ARep$ is searched to find entries whose AP combination contains the $S$ subset. When a match is found in cluster $C_k$:
    \begin{itemize}
        \item The frequency $freq(A_{k,s})$ is added to the score of cluster $C_k$,
        \item The contribution is weighted by the size $|S|$, so larger subsets contribute more.
    \end{itemize} 
    \item Finally, $A_{test}$ is assigned to the cluster $C_k$ with the highest score.
\end{enumerate}

The position estimation algorithm is then executed only within the selected cluster, thereby reducing search space and potentially improving accuracy and computational efficiency. 

The cluster selection process depends on the number of strongest APs considered ($N$), which is treated as a configurable hyperparameter of the proposed method.

\subsection{Position and floor estimation}
\label{subsec:position_estimation}
After a new fingerprint is assigned to a cluster, a learning algorithm is employed to estimate the corresponding position and floor. In the proposed method, the position estimation task is treated as a regression problem, while floor detection is handled as a classification problem. 

The specific requirements for these models depend on the clustering strategy employed (see Section~\ref{subsec:clustering_strategy}). When adopting the building-based strategy, clusters may contain fingerprints from multiple floors and, therefore, both position and floor estimation models are required. In contrast, when using the floor-based strategy, each cluster corresponds to a single floor, so the floor can be directly inferred from the assigned cluster. In this case, only the position estimation model is required.

The proposed method is flexible and agnostic to the choice of the learning algorithm, allowing any suitable model to be used for regression and classification tasks, such as KNN, gradient-boosted decision trees, or neural networks.

\section{Experimental setup}
\label{sec:4}
\begin{figure}[h]
    \centering
    \includegraphics[width=\linewidth, trim={.8cm .3cm .8cm .4cm}]{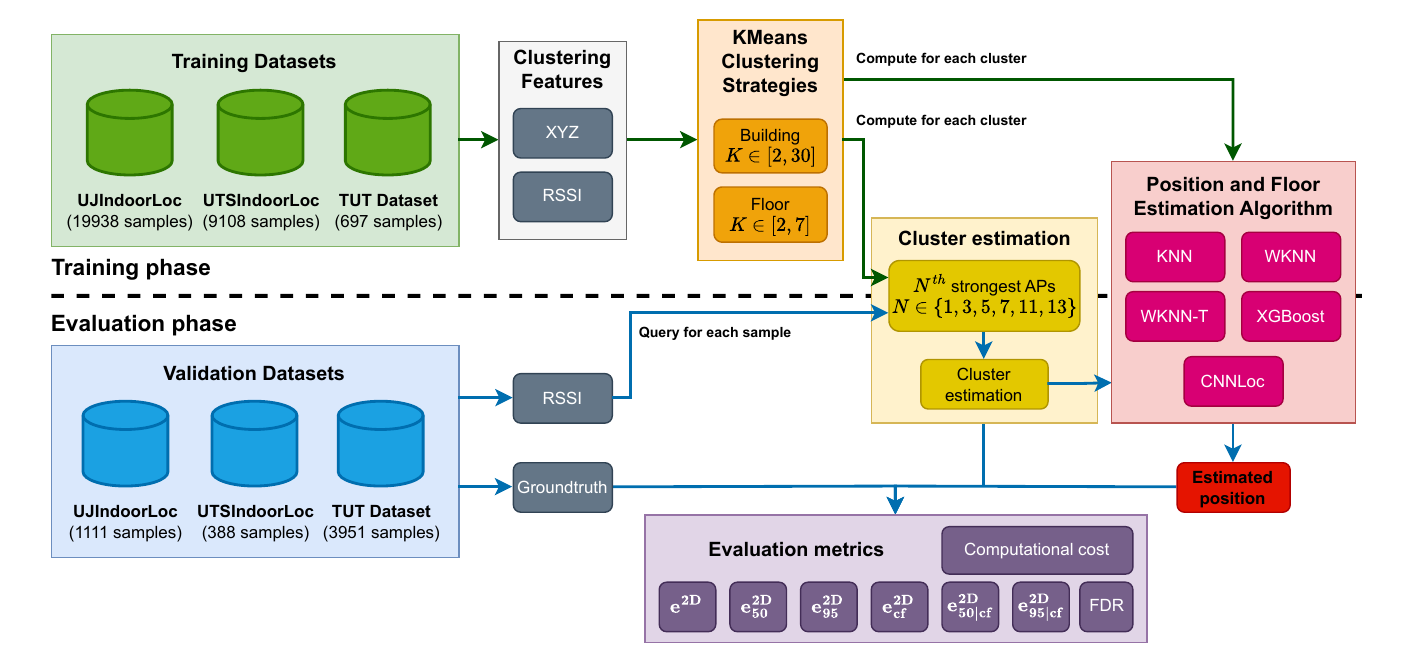}
    \caption{Experimental setup employed to evaluate the localisation and floor detection performance of the proposed method.}
    \label{fig:experimental_setup}
\end{figure}

This section describes the setup used to evaluate the proposed method across three public indoor positioning datasets. The evaluation procedure is depicted in Figure~\ref{fig:experimental_setup} and comprises the following components: a) the datasets employed (Section~\ref{subsec:datasets}); b) the hyperparameters used for the proposed method (Section~\ref{subsec:hyperparams}); c) the ML models applied for position and floor estimation (Section~\ref{subsec:ml}); d) the evaluation metrics to quantify the localisation and floor detection performance (Section~\ref{subsec:metrics}), and the computational environment used for the experiments (Section~\ref{subsec:env}).

\subsection{Datasets}
\label{subsec:datasets}
The performance of the proposed method was evaluated using three widely adopted and heterogeneous Wi-Fi fingerprinting datasets: UJIIndoorLoc~\cite{torres2014ujiindoorloc}, UTSIndoorLoc~\cite{song2019novel} and TUT~\cite{lohan2017}. These datasets differ significantly in terms of deployment environments, device diversity, and AP density, providing a robust basis for evaluating the method's generalisation. Each dataset is organised into training and testing subsets that contain the RSSI values recorded from all observable Wi-Fi APs at each sampled reference point.

Table \ref{tab:datasets} summarises their key properties, including the number of samples in each partition, the total number of APs, the variety and number of data-collection devices, the number of buildings and floors surveyed, and the spatial extent of the acquisition campaign. These complementary characteristics ensure that the evaluation reflects a wide range of realistic indoor localisation scenarios.

\begin{table}[]
    \centering
    \caption{Comparison of the main characteristics of the Wi-Fi fingerprinting datasets used in the evaluation.}
    \begin{tabular}{lccccccc}
    \toprule
        Dataset      & APs & Training samples & Testing samples & Devices & Buildings & Floors & Area \\ \midrule
        UJIIndoorLoc~\cite{torres2014ujiindoorloc} & 520 & 19938 & 1111 & 25 & 3 & 4/4/5 & \SI{110000}{\meter^2}\\
        UTSIndoorLoc~\cite{song2019novel} & 589 & 9108  & 388  & 1  & 1 & 16 & \SI{44000}{\meter^2} \\
        TUT~\cite{lohan2017}          & 992 & 697   & 3951 & 21 & 1 & 5 & \SI{22570}{\meter^2} \\
    \bottomrule
    \end{tabular}
    \label{tab:datasets}
\end{table}

\subsection{Method hyperparameters}
\label{subsec:hyperparams}

\subsubsection{Number of clusters: $K$}
For both building-based and floor-based clustering strategies, multiple values of the number of clusters $K$ (not to be confused with the number of neighbours used in KNN-based models) were evaluated to assess the sensitivity of the proposed method to cluster granularity. Specifically, values of $K \in [2, 30]$ were considered for building-based clustering in the case of UTSIndoorLoc and TUT datasets, while $K \in [2, 16]$ was used independently for each of the three buildings in UJIIndoorLoc, due to computational limitations. For the floor-based clustering strategy, values of $K \in [2,7]$ were used for floor-based clustering due to the smaller spatial extent and data availability at the floor level.

\subsubsection{Number of APs: $N$}
The number of strongest APs ($N$) used to assign fingerprints to clusters was varied in the experiments, with $N \in [1,5]$. This range of values was selected since a small number of dominant APs within each cluster can capture most location-dependent information, while limiting the computational complexity.

\subsection{Machine learning algorithms}
\label{subsec:ml}
We selected the KNN, XGBoost~\cite{chen2016xgboost} and CNNLoc~\cite{song2019novel} algorithms for position and floor estimation to evaluate the proposed method on the three public datasets. 

\subsubsection{$k$-Nearest Neighbours}
The KNN algorithm computes the similarity of a given testing fingerprint to all the fingerprints in the training dataset. To compute the similarity between fingerprints, we employ the well-known Euclidean distance, defined as:

\begin{equation}
    d(\vec{x}, \vec{y}) = \sqrt{\sum_{n=1}^N{(x_{APn}-y_{APn})^2}}\,,
\end{equation}

where $\vec{x}$ and $\vec{y}$ are the corresponding RSSI vector of two fingerprints. The position and floor of the testing fingerprint are then estimated based on the average coordinates (i.e., regression) and the most frequent floor (i.e., classification) of the $k$ most similar fingerprints. Since several approaches can be used to perform the estimations, we employed three KNN variants:

\begin{itemize}
    \item \textbf{Baseline KNN}: the estimated coordinates are computed as the arithmetic mean of the coordinates of the $k$ nearest neighbours, defined as:
    \begin{equation}
        \hat{y} = \frac{1}{k} \sum_{i\in \mathcal{N}_k} y_i \,,
    \end{equation}
    where $\mathcal{N}_k$ are the $k$ nearest fingerprints of the testing fingerprint. The estimated floor is the most frequent floor among the $k$ nearest neighbours:
    \begin{equation}
        \hat{y} = \operatorname*{mode}_{i \in \mathcal{N}_k} (y_i)\,.
    \end{equation}
    \item \textbf{Weighted KNN (WKNN)}: given the distance $d_i$ between the testing fingerprint and the neighbour $i$, weights are defined as the inverse distance:
    \begin{equation}
        w_i = \frac{1}{d_i}.
    \end{equation}

    Then, the estimated coordinates are the weighted average of the coordinates of the $k$ nearest neighbours, and the predicted floor is the floor with the maximum sum of weights, as defined by:
    \begin{equation}
    \hat{y} = 
    \frac{\sum_{i \in \mathcal{N}_k} w_i \, y_i}{\sum_{i \in \mathcal{N}_k}\, w_i} \,,
    \end{equation}
    \begin{equation}
        \hat{y} = \operatorname*{arg\,max}_c \sum_{i \in \mathcal{N}_k} w_i \, I(y_i = c) \,.
    \end{equation}
    \item \textbf{WKNN with tie resolution (WKNN-T)}: follows the same procedure as WKNN, but additionally includes all neighbours whose distance equals that of the $k$-th nearest neighbour, ensuring that distance ties are consistently handled during estimation.
\end{itemize}

The values used for the hyperparameter $k$ during the evaluation were $k \in {1, 3, 5, 7, 9, 11, 13}$.

\subsubsection{XGBoost}
XGBoost is an optimised implementation of gradient boosted decision trees, designed to achieve high predictive performance and computational efficiency. The algorithm builds an ensemble of regression trees, where each tree is trained to fix the residual errors made by the previously trained trees, thereby minimising the specified loss function~\cite{chen2016xgboost}. 

This model supports both regression and classification tasks, corresponding to location and floor prediction, respectively. During the evaluation, all hyperparameters were kept at their default settings, except for the maximum number of boosted decision trees. This parameter was varied using $n \in {50, 100, 150, 200}$.

\subsubsection{CNNLoc}
CNNLoc is a convolutional neural network (CNN) architecture specifically designed for indoor positioning using Wi-Fi fingerprinting~\cite{song2019cnnloc, song2019novel}. The method treats the RSSI fingerprint as a 2D representation, enabling the exploitation of local and spatial correlations in the signal space through convolutions. The CNNLoc is composed of a Stacked Auto-Encoder (SAE) and a 3-layered 1D CNN, and supports both location estimation (i.e., regression) and floor detection (i.e., classification) tasks. In this work, the CNNLoc implementation provided by the original authors was adopted as a baseline, but several hyperparameters adjustments were introduced to improve its learning capabilities within the proposed method. Table~\ref{tab:cnnloc-setup} summarises the architecture details and training hyperparameters used for each CNNLoc variant.

\begin{table}[]
    \caption{CNNLoc architecture and training hyperparameters used for location (regression) and floor (classification) estimation. Acronyms: ReLU (Rectified Linear Unit)~\cite{agarap2018deep}, MSE (Mean Squared Error), ELU (Exponential Linear Unit)~\cite{clevert2015fast}, Adam (Adaptive Moment Estimation)~\cite{kingma2014adam}, Adamax (Adam variant)~\cite{kingma2014adam}.}
    \centering
    \begin{tabular}{lcc}
    \toprule
        Parameter                   & Location model & Floor model \\
    \midrule                                
        SAE hidden layers           & (128,64,128)    & (128,64,128)  \\ 
        SAE optimizer               & Adam(lr=0.0001) & Adam(lr=0.0001) \\
        SAE activation function     & ReLU            & ReLU           \\
        SAE loss                    & MSE             & MSE           \\
        1D CNN filters              & (99,66,33)      & (99,66,33) \\
        1D CNN kernel size          & 1$\times$22     & 1$\times$22   \\
        1D CNN optimizer            & Adamax(lr=0.005)& Adamax(lr=0.005) \\
        1D CNN activation function  & ELU             & ELU \\
        1D CNN loss                 & MSE             & MSE \\
        Output activation function  & ELU             & Softmax \\
        Batch Size                  & 66              & 66 \\
        Training Epochs             & 60              & 40 \\
    \bottomrule
    \end{tabular}
   
    \label{tab:cnnloc-setup}
\end{table}

\subsection{Evaluation metrics}
\label{subsec:metrics}
To assess the performance of the proposed method, two complementary aspects are evaluated: a) floor identification accuracy and b) horizontal localisation in the two-dimensional ($x$-$y$) plane. Therefore, the primary metrics used in this study are the Floor Detection Rate (FDR), the overall 2D horizontal positioning error ($e^{2D}$), and the 2D horizontal error conditioned on correct floor identification ($e^{2D}_{cf}$).

Floor identification accuracy is quantified using the FDR, defined as the proportion of samples for which the estimated floor matches the ground-truth floor. This metric captures the method's ability to accurately determine the user's vertical position in multi-floor environments.

Horizontal localisation accuracy is evaluated using the Euclidean distance between the estimated and ground-truth coordinates on the $x$–$y$ plane. In addition, to separate horizontal localisation accuracy from vertical misclassification effects and to provide insight into the method's intrinsic horizontal accuracy, we also compute the 2D positioning error conditioned on correct floors identification, excluding samples for which the floor was incorrectly predicted. For both metrics, summary statistics are reported, including the mean error and $50^{th}$ ($e^{2D}_{50}$ and $e^{2D}_{50|cf}$) and $95^{th}$ ($e^{2D}_{95}$ and $e^{2D}_{95|cf}$) percentiles.

In addition, the computational cost of the cluster creation and estimation procedure described in Section~\ref{subsec:cluster_estimation} is empirically evaluated by measuring the required computation time. The effect of the proposed method in its remaining components (i.e., K-Means, model training, and model inference) is discussed descriptively, based on well-established computational properties of the corresponding algorithms.

\subsection{Computational environment}\label{subsec:env}
All experiments were implemented in Python 3.11.11 using standard scientific and ML libraries, including NumPy v2.1.3, Pandas v2.2.3, scikit-learn v1.6.1, XGBoost v3.0.2 and TensorFlow v2.19.0. The experiments were executed on a computer running Ubuntu 24.04.01 equipped with an AMD Ryzen 5 7600X CPU and \SI{64}{\giga\byte} of RAM. The data and code to reproduce the results presented in the following section are available in Zenodo~\cite{matey2026zenodo}.

\section{Results}
\label{sec:5}
This section presents the results of applying the proposed method under the scenarios described in the previous section. The following subsections focus on the positioning performance obtained by the proposed method under each scenario (Section~\ref{subsec:pos_performance}), an analysis of the impact of hyperparameters $N$ (Section~\ref{subsec:analysis_n}) and $K$ (Section~\ref{subsec:analysis_k}) on the positioning errors, a computational cost analysis of the proposed method (Section~\ref{subsec:comp_cost}) and a comparison with the results reported in other state-of-the-art methods (Section~\ref{subsec:comparison}).

\subsection{Overall positioning performance}\label{subsec:pos_performance}

\begin{figure}[h]
    \centering
    \begin{subfigure}[b]{0.32\textwidth}
        \includegraphics[width=\linewidth, trim={0 1cm 2cm 2.2cm}, clip]{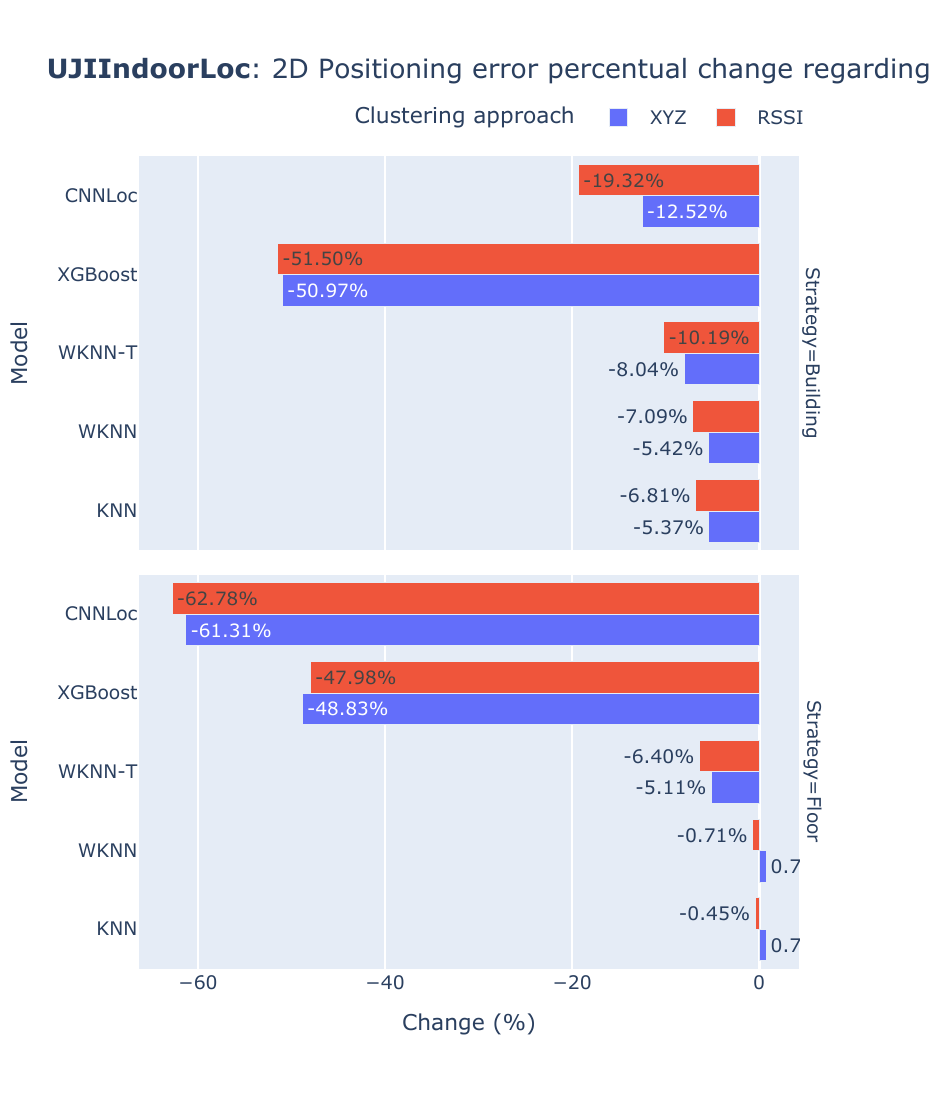}
        \caption{UJIIndoorLoc.}
        \label{subfig:uji_error_pct}
    \end{subfigure}
    \begin{subfigure}[b]{0.32\textwidth}
        \includegraphics[width=\linewidth, trim={0cm 1cm 2cm 2.2cm}, clip]{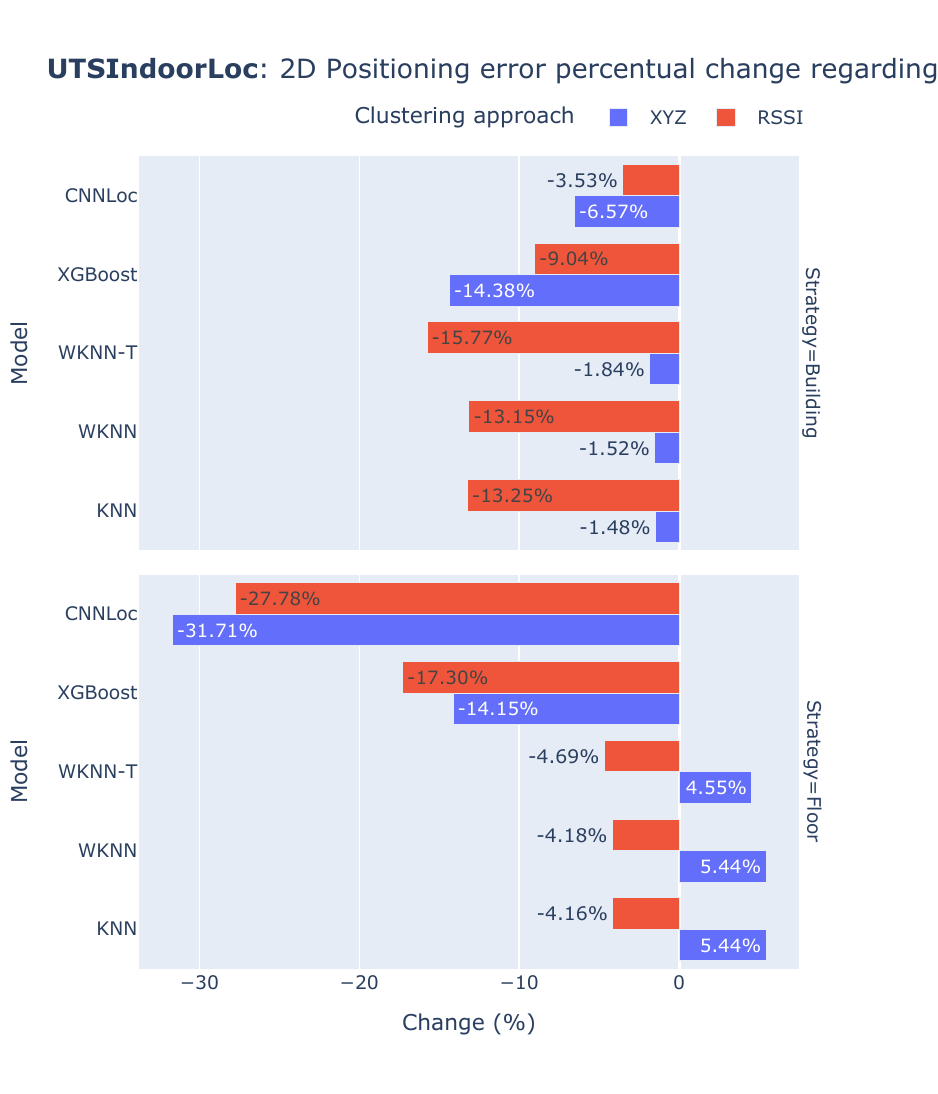}
        \caption{UTSIndoorLoc.}
        \label{subfig:uts_error_pct}
    \end{subfigure}
    \begin{subfigure}[b]{0.32\textwidth}
        \includegraphics[width=\linewidth, trim={0cm 1cm 2cm 1.8cm}, clip]{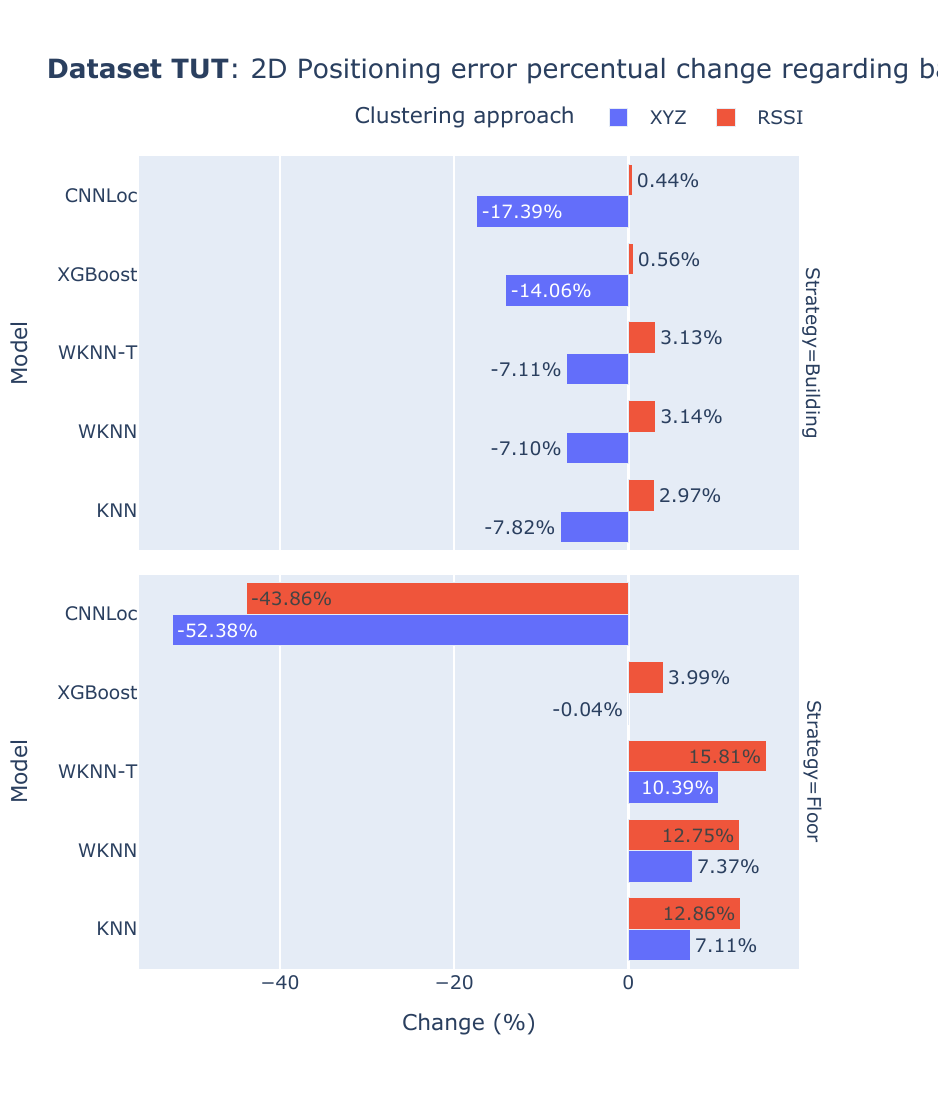}
        \caption{TUT.}
        \label{subfig:tut_error_pct}
    \end{subfigure}
    \caption{2D positioning error percentage change of clustering approaches compared with the baseline approach in each dataset. A \textbf{negative increment} indicates that the positioning error \textbf{has been improved} compared with the baseline. Top figures are for the building strategy and bottom figures for floor strategy. Blue and red bars represent percentual change of XYZ clustering vs. Baseline and RSSI clustering vs Baseline, respectively.}
    \label{fig:error_pct}
\end{figure}

\begin{figure}[h]
    \centering
    \begin{subfigure}[b]{0.32\textwidth}
        \includegraphics[width=\linewidth, trim={0 1cm 2cm 2.2cm}, clip]{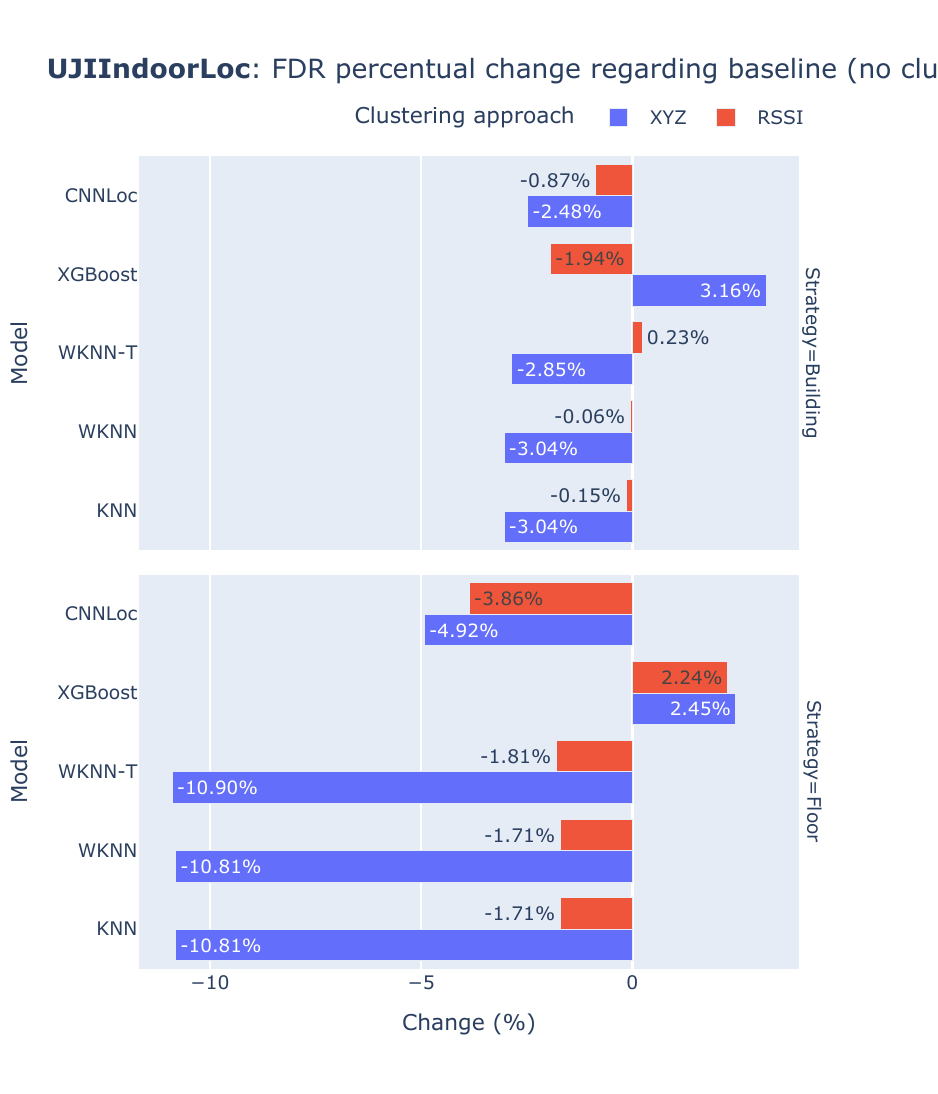}
        \caption{UJIIndoorLoc.}
        \label{subfig:uji_fdr_pct}
    \end{subfigure}
    \begin{subfigure}[b]{0.32\textwidth}
        \includegraphics[width=\linewidth, trim={0cm 1cm 2cm 2.2cm}, clip]{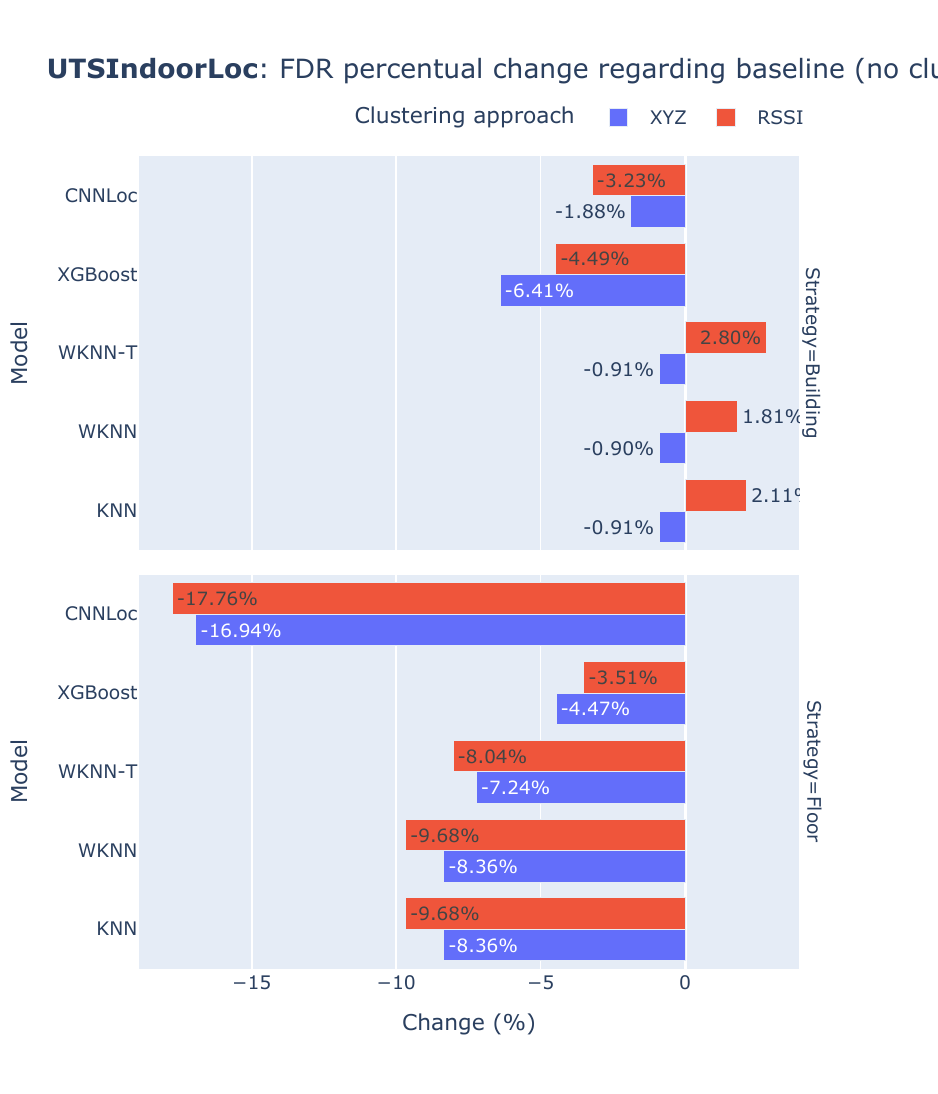}
        \caption{UTSIndoorLoc.}
        \label{subfig:uts_fdr_pct}
    \end{subfigure}
    \begin{subfigure}[b]{0.32\textwidth}
        \includegraphics[width=\linewidth, trim={0cm 1cm 2cm 1.8cm}, clip]{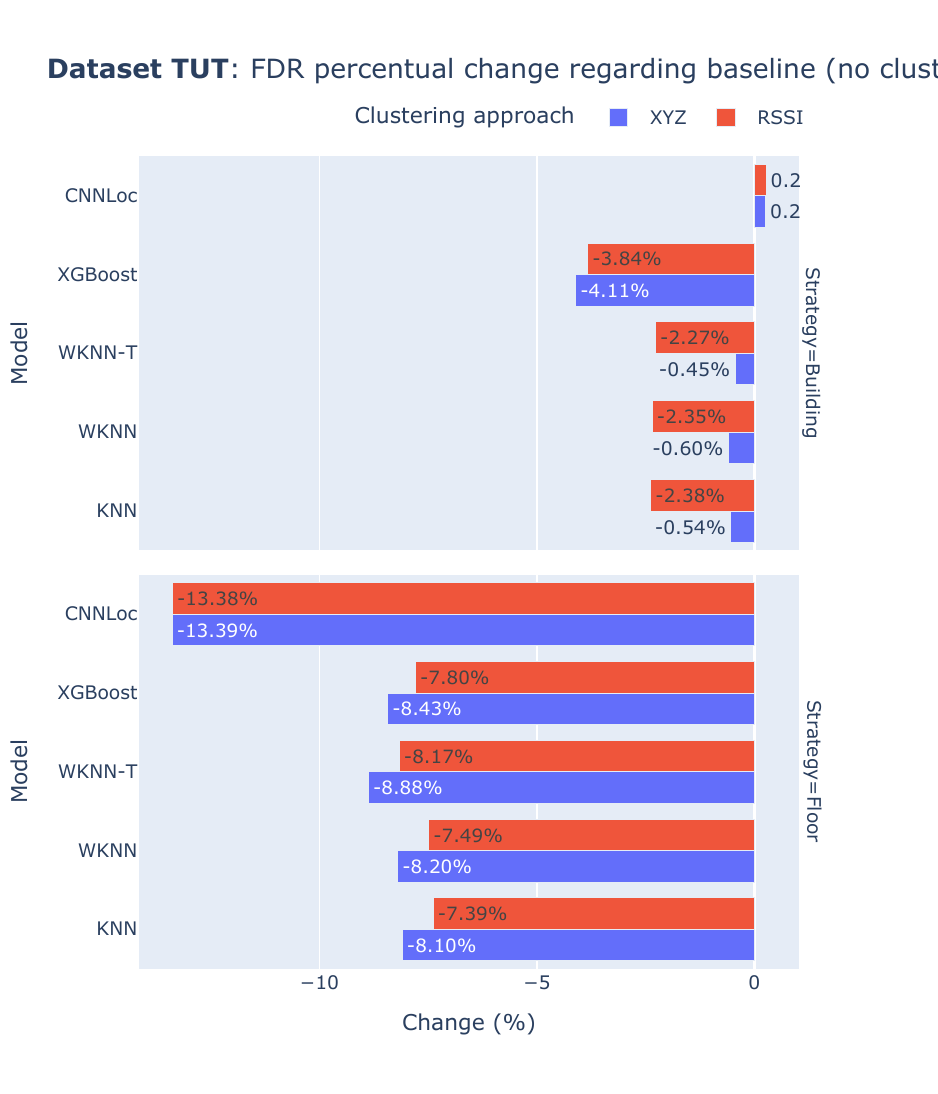}
        \caption{TUT.}
        \label{subfig:tut_fdr_pct}
    \end{subfigure}
    \caption{Floor detection rate percentage change of clustering approaches compared with the baseline approach in each dataset. A \textbf{positive increment} indicates that the FDR \textbf{has been improved} compared with the baseline. Top figures are for the building strategy and bottom figures for floor strategy. Blue and red bars represent percentual change of XYZ clustering vs. Baseline and RSSI clustering vs Baseline, respectively.}
    \label{fig:fdr_pct}
\end{figure}

Tables~\ref{tab:uji}, \ref{tab:uts} and \ref{tab:tut} show the best positioning errors and FDR values for the UJIIndoorLoc, UTSIndoorLoc and TUT datasets, respectively, and each selected ML model, clustering approach and clustering strategy. Additionally, Figures~\ref{fig:error_pct} and \ref{fig:fdr_pct} show the percentage change in 2D positioning error and FDR of the clustering approaches compared with the baseline (i.e., without clusters) for each strategy and ML model. In the following subsections, a comparison is presented to identify which clustering approach (Section~\ref{subsec:res:clustering_aproach}), clustering strategy (Section~\ref{subsec:res:strategy}) and learning model (Section~\ref{subsec:res:models}) provides the best results across each dataset.
\begin{table}[h]
    \scriptsize
    \tabcolsep 1.5pt
    \centering
    \caption{Best positioning errors (\unit{\meter}) and FDR (\unit{\percent}) in the UJIIndoorLoc dataset. In the clustering approaches, $N$ is the number of APs considered for cluster estimation and $K$ is the number of clusters.}
    \resizebox{\linewidth}{!}{\input{tables/uji}}
    \label{tab:uji}
\end{table}

\begin{table}[h]
    \scriptsize
    \tabcolsep 1.5pt
    \centering
    \caption{Best positioning errors (\unit{\meter}) and FDR (\unit{\percent}) in the UTSIndoorLoc dataset. In the clustering approaches, $N$ is the number of APs considered for cluster estimation and $K$ is the number of clusters.}
    \resizebox{\linewidth}{!}{\input{tables/uts}}
    \label{tab:uts}
\end{table}

\begin{table}[h]
    \scriptsize
    \tabcolsep 1.5pt
    \centering
    \caption{Best positioning errors (\unit{\meter}) and FDR (\unit{\percent}) in the TUT dataset. In the clustering approaches, $N$ is the number of APs considered for cluster estimation and $K$ is the number of clusters.}
    \resizebox{\linewidth}{!}{\input{tables/tut}}
    \label{tab:tut}
\end{table}

\subsubsection{Clustering approach comparison}
\label{subsec:res:clustering_aproach}
In the UJIIndoorLoc dataset (Table~\ref{tab:uji}), the RSSI-based clustering approach obtains the best overall results in $e^{2D}$, $e^{2D}_{cf}$, and their $95^{th}$ percentiles when using the building-based strategy, while the baseline approach achieves lower $50^{th}$ percentile errors and higher FDR. For the floor-based strategy, the baseline approach achieves the best performance for the KNN-based models, except for the $e^{2D}$, which is slightly lower when RSSI-based clustering is applied. However, the baseline approach performs poorly with XGBoost and CNNLoc, where both XYZ- and RSSI-based clustering approaches yield better results. 

Regarding the UTSIndoorLoc dataset (Table~\ref{tab:uts}), the RSSI-based clustering approach again obtains the best overall results when using both building- and floor-based strategies, except with the XGBoost and CNNLoc, where XYZ-based clustering performs better. It is noticeable that, when using the floor-based strategy, the baseline method achieves substantially higher FDR values compared with clustering-based approaches.

Finally, for the TUT dataset (Table~\ref{tab:tut}), the XYZ-based clustering approach achieves the best error metrics in the building strategy, although the baseline method obtains the highest FDR rates. When using the floor-based strategy, the baseline method performs best with the KNN-based models, while the XYZ-based clustering yields better performance with the XGBoost and CNNLoc, except for the FDR metric. 

As shown in Figure~\ref{fig:error_pct}, the improvement achieved by the clustering approaches depends on both the dataset and the clustering strategy: 

\begin{itemize}
    \item In UJIIndoorLoc, both clustering approaches reduce the $e^{2D}$ compared to the baseline approach in all evaluated combinations.
    \item In UTSIndoorLoc, RSSI-based clustering consistently improves the positioning errors in all scenarios, while XYZ-based clustering worsens the baseline method in the KNN-based algorithms under the floor-based strategy.
    \item In TUT, XYZ-based clustering improves positioning errors under the building-based strategy, while RSSI-based clustering presents worse results than the baseline in all scenarios except for CNNLoc under the floor-based strategy. 
\end{itemize}
 
While both clustering approaches tend to reduce the FDR, their behaviour also depends on the dataset, as shown in Figure~\ref{fig:fdr_pct}. For instance, RSSI-based clustering reduces FDR less than XYZ in UJIIndoorLoc, but more than in UTSIndoorLoc. 

Overall, both clustering approaches tend to provide reduced mean errors and also the lowest $95^{th}$ percentile errors. However, these improvements are obtained at the cost of reduced FDR, with the baseline method systematically yielding the best results in this metric.

\subsubsection{Strategy comparison: building vs floor}\label{subsec:res:strategy}
Regarding the comparison between clustering strategies across all datasets --UJIIndoorLoc (Table~\ref{tab:uji}), UTSIndoorLoc (Table~\ref{tab:uts}) and TUT (Table~\ref{tab:tut})--, a consistent pattern emerges. For models using XYZ- or RSSI-based clustering, the building-based strategy generally outperforms the floor-based strategy, particularly reducing the $95^{th}$ percentile errors, which correspond to the largest positioning deviations. 
In the baseline models, both strategies often yield similar performance metrics in the KNN-based models, but CNNLoc provides significantly worse results with the floor strategy.

In addition, the building-based strategy also consistently achieves higher FDR values, highlighting its advantage in floor prediction. The only exceptions are the baseline models of CNNLoc in UJIIndoorLoc and TUT, and some of the clustering XGBoost models in UJIIndoorLoc and UTSIndoorLoc.

Overall, these results indicate that adopting a building-level training strategy combined with XYZ- or RSSI-based clustering provides the best trade-off between horizontal accuracy and floor detection across datasets and learning models.

\subsubsection{Model-specific observations}
\label{subsec:res:models}
For the KNN-based models (KNN, WKNN, and WKNN-T), the impact of clustering strategies varies across datasets. In UJIIndorLoc, under the building-based strategy, both XYZ- and RSSI-based clustering generally improve baseline positioning errors, with RSSI-based clustering achieving the lowest overall 2D errors, although the $50^{th}$ percentiles and FDR remain slightly higher in the baseline approach. With the floor-based strategy, baseline models often outperform the clustering-based configurations. In UTSIndoorLoc, RSSI-based clustering provides the best results for both strategies, except for the $95^{th}$ percentile errors and FDR under the floor-based strategy. For TUT, results differ between strategies: XYZ-based clustering reduces errors relative to the baseline under the building-based strategy, while clustering presents no improvement under the floor-based strategy. Overall, KNN-based models benefit most from clustering when combined with the building-level training strategy.

XGBoost shows consistent improvement with clustering across all datasets. In UJIIndoorLoc, RSSI-based clustering minimises most positioning errors under the building-based strategy, while XYZ-based clustering slightly outperforms RSSI in terms of the $50^{th}$ percentiles and FDR. Under the floor-based strategy, XYZ-based clustering generally achieves the best performance. For UTSIndoorLoc (building-based strategy only) and TUT, XYZ-based clustering yields the lowest positioning errors for both strategies. Importantly, in these datasets, both clustering approaches improve FDR relative to the baseline, highlighting a combined benefit in horizontal position estimation and floor detection performance.

CNNLoc shows substantial performance gains when combined with clustering. In UJIIndoorLoc, RSSI-based clustering provides the lowest positioning errors under the building-based strategy, while XYZ-based clustering performs best under the floor-based strategy. In UTSIndoorLoc and TUT, XYZ-based clustering consistently delivers the largest reductions in positioning errors. Across all datasets, the highest FDR values are obtained by the baseline configurations, indicating that while clustering improves horizontal position estimation, it does not enhance floor detection performance for CNNLoc.

\subsection{Hyperparameter $N$ impact analysis}
\label{subsec:analysis_n}
\begin{figure}[h]
    \centering
    \begin{subfigure}[b]{0.32\textwidth}
        \includegraphics[width=\linewidth, trim={0.8cm 1cm 2cm 1.5cm}, clip]{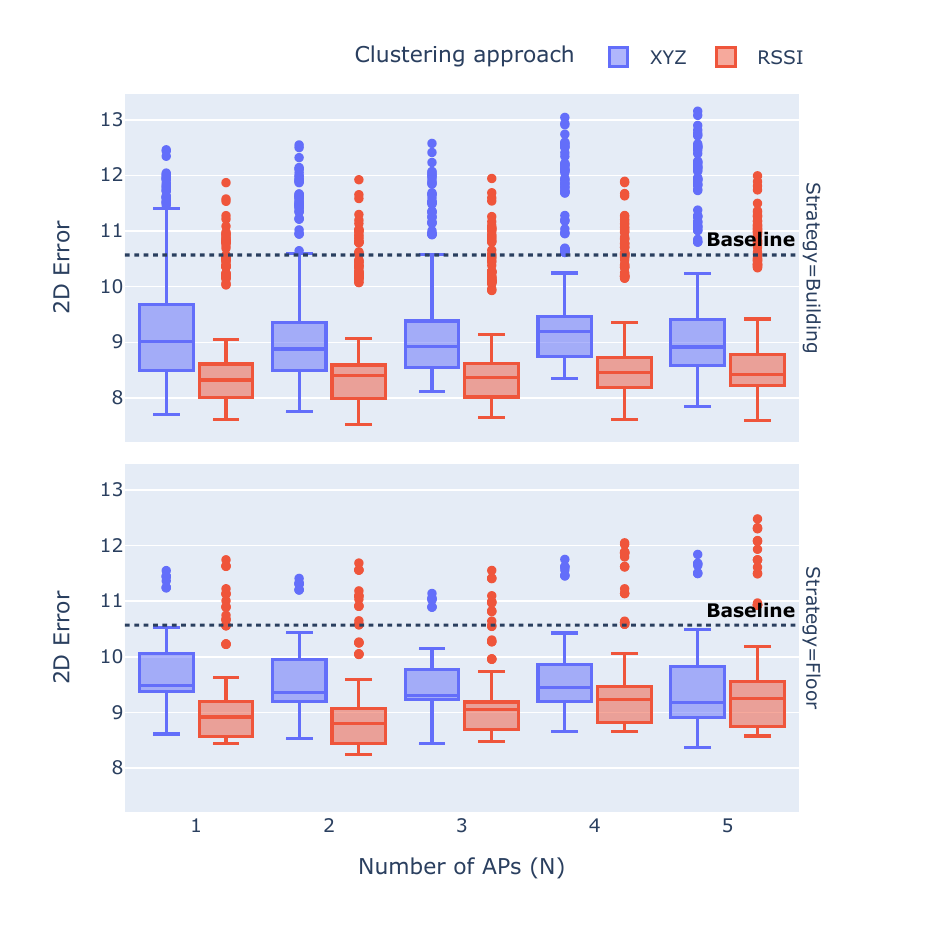}
        \caption{UJIIndoorLoc.}
        \label{subfig:uji_n_boxplot}
    \end{subfigure}
    \begin{subfigure}[b]{0.32\textwidth}
        \includegraphics[width=\linewidth, trim={0.8cm 1cm 2cm 1.5cm}, clip]{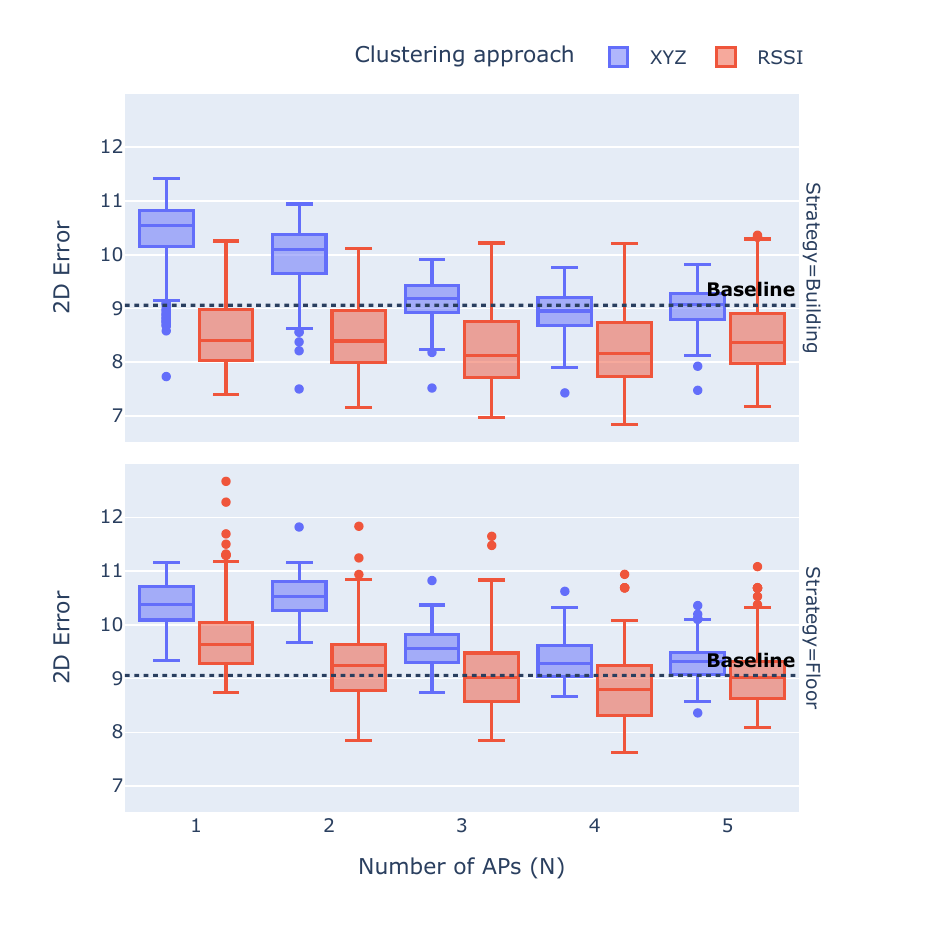}
        \caption{UTSIndoorLoc.}
        \label{subfig:uts_n_boxplot}
    \end{subfigure}
    \begin{subfigure}[b]{0.32\textwidth}
        \includegraphics[width=\linewidth, trim={{0.8cm 1cm 2cm 0.75cm}}, clip]{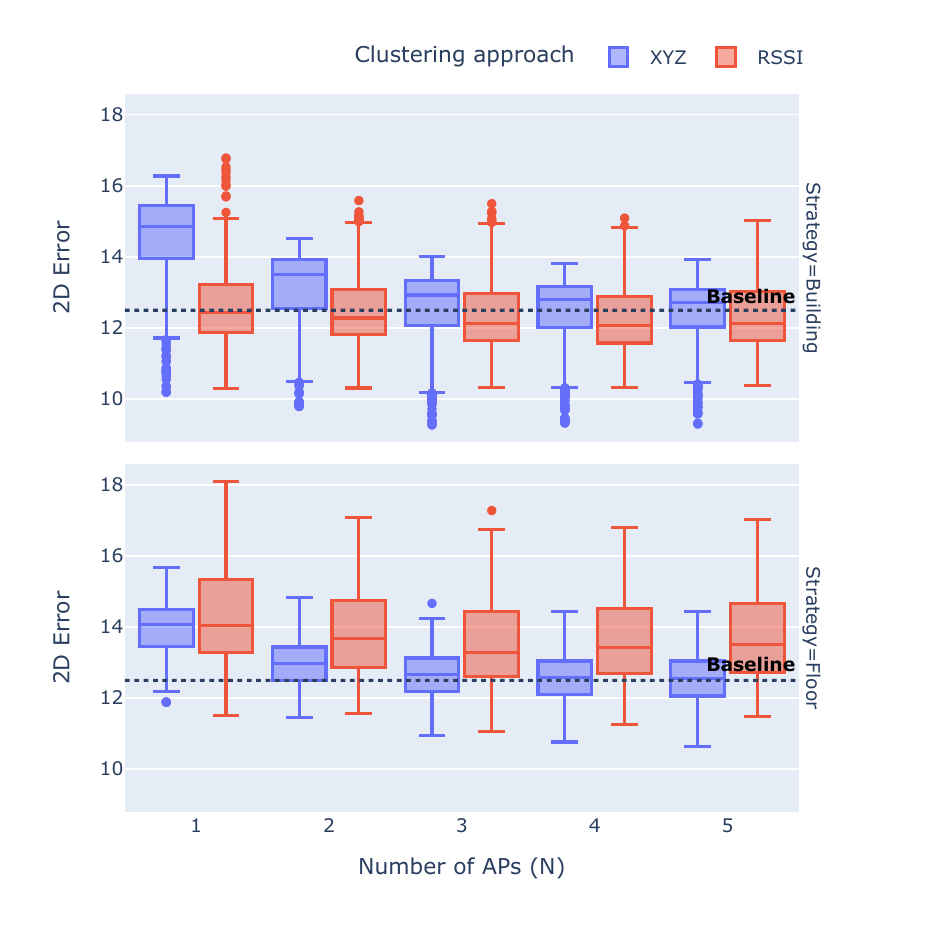}
        \caption{TUT.}
        \label{subfig:tut_n_boxplot}
    \end{subfigure}
    \caption{Change of positioning error regarding the number of selected APs ($N$). Errors from all models and their configurations were aggregated. The upper subfigures correspond to the building-based strategy, whereas the lower subfigures correspond to the floor-based strategy. Blue and red boxes represent error distribution (i.e., boxplots) for XYZ-based and RSSI-based clustering, respectively; black dashed lines indicate the baseline aggregated mean errors.}
    \label{fig:n_boxplots}
\end{figure}

Figure~\ref{fig:n_boxplots} illustrates the impact of the hyperparameter $N$ (number of APs) in the proposed method on the 2D positioning error ($e^{2D}$) across all evaluated models and datasets. For the UJIIndoorLoc (Figure~\ref{subfig:uji_n_boxplot}), varying $N$ does not substantially affect the median positioning error when the building-based strategy is used, indicating that increasing the number of APs beyond a minimal subset does not necessarily translate into improved accuracy in this scenario. Nevertheless, the reduction in the whisker range observed for higher values of $N$ under XYZ-based clustering suggests that larger $N$ values can improve the stability of the method by limiting extreme errors, even when the median performance remains unchanged. Under the floor-based strategy, error variability remains largely insensitive to $N$, while the median error shows opposite trends depending on the clustering approach (improvements in XYZ-based and decrements in RSSI-based clustering), implying that the choice of $N$ may interact with model characteristics rather than producing uniform performance gains.

Similar trends are observed for the UTSIndoorLoc (Figure~\ref{subfig:uts_n_boxplot}) and TUT (Figure~\ref{subfig:tut_n_boxplot}) datasets. In these datasets, increasing $N$ from $1$ to $3$ yields the most tangible benefits, with a clear reduction in positioning error under both strategies for XYZ-based clustering and under floor-based strategy for RSSI-based clustering. However, beyond $N=3$, the positioning error distributions stabilise, and further increases in $N$ provide limited or no additional improvements in either positioning accuracy or robustness. This behaviour indicates that a moderate number of APs is sufficient to capture relevant information for positioning, and that selecting larger values of $N$ may unnecessarily increase system complexity without a proportional performance gain.

Overall, the analysis indicates that $N$ should be selected conservatively, as most of the achievable gains in positioning accuracy are obtained with a small number of APs. In particular, increasing $N$ from $1$ to $3$ yields the most consistent performance gains across datasets, while larger values provide only limited benefits. Therefore, choosing $N\approx3$ represents a good trade-off between localisation accuracy and computational complexity.

\subsection{Hyperparameter $K$ impact analysis}
\label{subsec:analysis_k}
\begin{figure}
    \centering
    \begin{subfigure}[b]{0.32\textwidth}
        \includegraphics[width=\linewidth, trim={0.8cm 0.8cm 2cm 1.5cm}, clip]{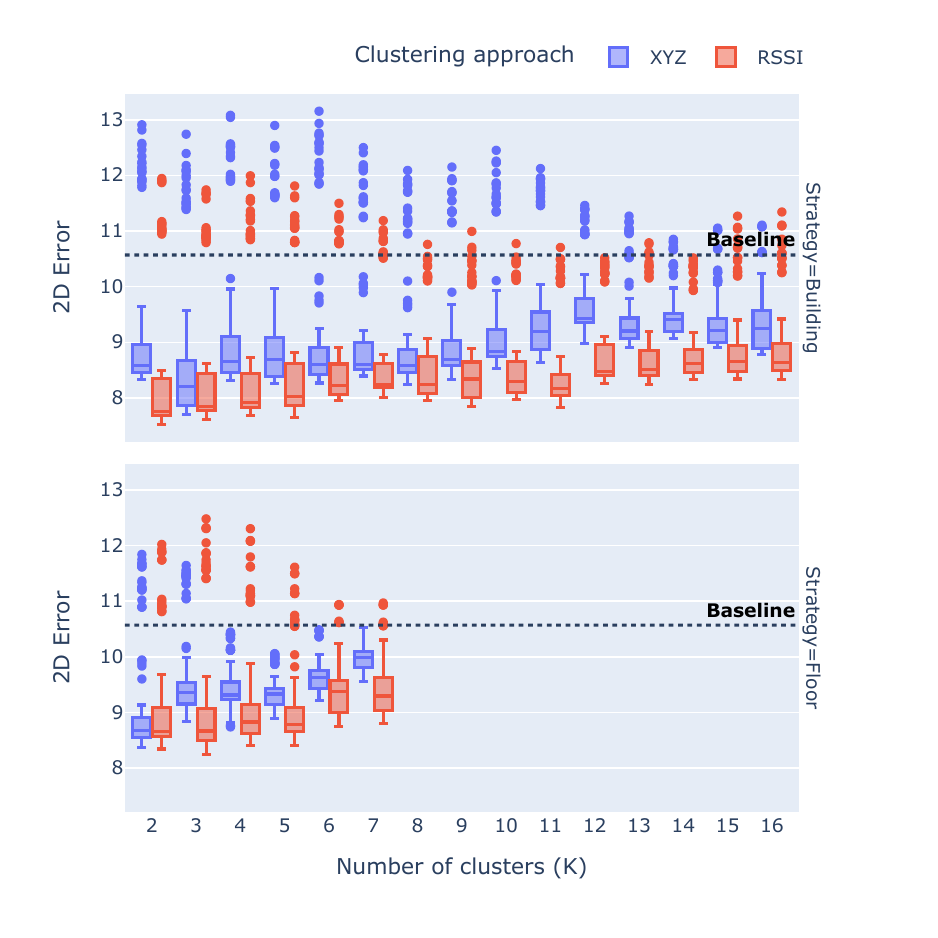}
        \caption{UJIIndoorLoc.}
        \label{subfig:uji_k_boxplot}
    \end{subfigure}
    \begin{subfigure}[b]{0.32\textwidth}
        \includegraphics[width=\linewidth, trim={0.8cm 0.8cm 2cm 1.5cm}, clip]{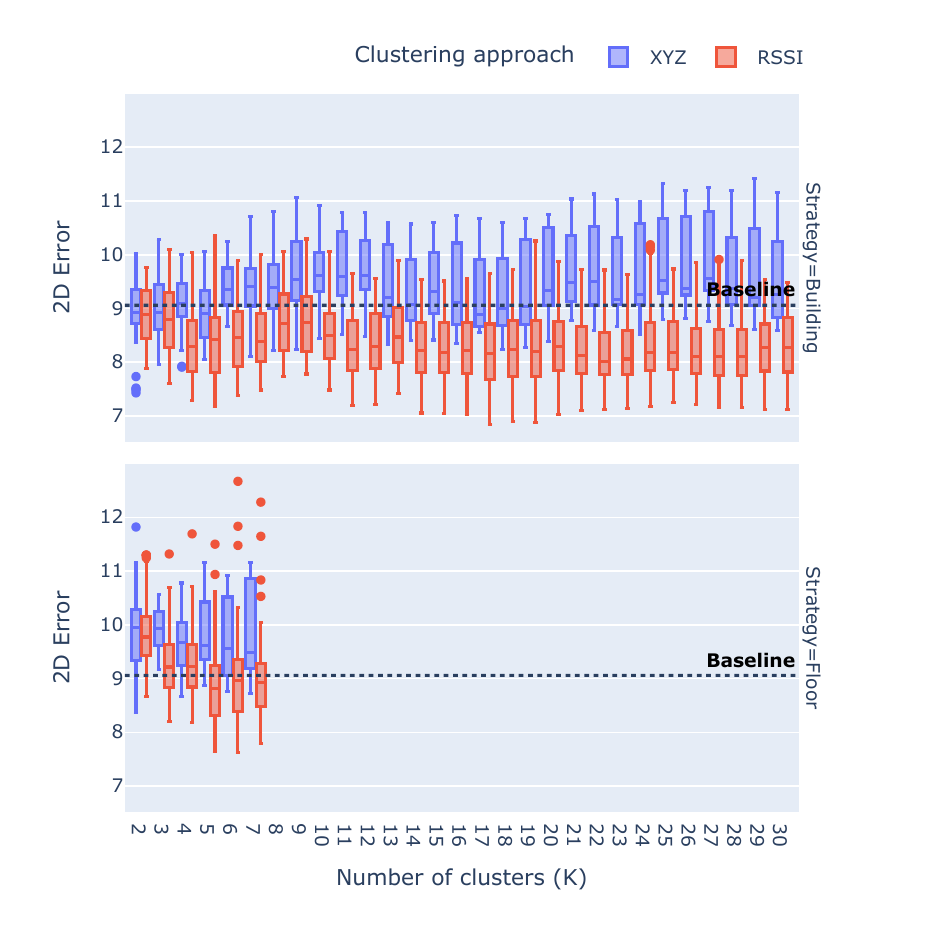}
        \caption{UTSIndoorLoc.}
        \label{subfig:uts_k_boxplot}
    \end{subfigure}
    \begin{subfigure}[b]{0.32\textwidth}
        \includegraphics[width=\linewidth, trim={{0.8cm 0.8cm 2cm 0.75cm}}, clip]{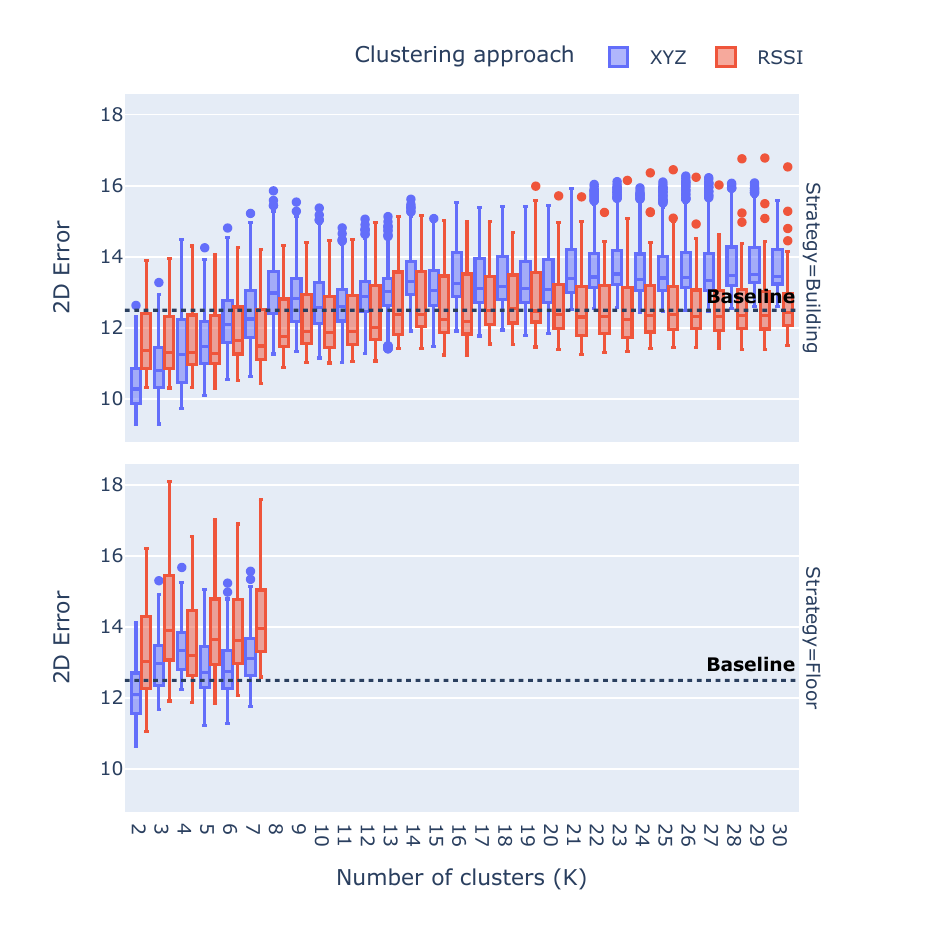}
        \caption{TUT.}
        \label{subfig:tut_k_boxplot}
    \end{subfigure}
    \caption{Change of positioning error as a function of the number of computed clusters ($K$). Errors from all models and their configurations were aggregated. The upper subfigures correspond to the building-based strategy, whereas the lower subfigures for floor-based strategy. Blue and red boxes represent error distribution (i.e., boxplots) XYZ- and RSSI-based clustering, respectively; black dashed lines indicate the baseline aggregated mean errors.}
    \label{fig:k_boxplots}
\end{figure}

Figure~\ref{fig:k_boxplots} illustrates the effect of the hyperparameter $K$ (number of clusters) in the proposed method on the 2D positioning error ($e^{2D}$) across all evaluated models and datasets. For the UJIIndoorLoc dataset (Figure~\ref{subfig:uji_k_boxplot}), the lowest positioning errors are achieved with a small number of clusters. Specifically, optimal performance is observed for $K=2$ or $K=3$, depending on the clustering approach and strategy. Increasing $K$ beyond these values generally results in a gradual degradation of positioning accuracy, indicating that excessive data partitioning can be detrimental. This behaviour suggests that, for UJIIndoorLoc, coarse clustering is sufficient to capture the dominant spatial and radio characteristics of the environment, and that selecting larger values of $K$ may introduce unnecessary fragmentation without improving localisation performance, even though improvements over the baseline may still be observed.

A different behaviour is observed for the UTSIndoorLoc dataset (Figure~\ref{subfig:uts_k_boxplot}). Under the building-based strategy, both clustering approaches exhibit an oscillatory response as $K$ increases, with performance alternating between local improvements and degradations. In this case, the optimal values of $K$ are not concentrated at small cluster values, but rather at higher number of clusters, with best positioning errors occurring at intermediate ($K=17$ for RSSI-based clustering) to large ($K=22$ for XYZ-based clustering) values of $K$. This variability indicates a higher sensitivity of the method to the choice of $K$, suggesting that the UTSIndoorLoc radio environment is more heterogeneous and benefits from finer-grained partitioning. Under the floor-based strategy, median errors generally decrease as $K$ increases for both clustering approaches, although the minimum error for XYZ-based clustering is still achieved at $K=1$. These results imply that increasing $K$ can be beneficial, but the choice of an optimal value remains non-trivial and configuration-dependent.

The TUT dataset (Figure~\ref{subfig:tut_k_boxplot}) exhibits a trend closer to UJIIndoorLoc, where the best performance is again obtained with a small number of clusters. Specifically, optimal results are observed at $K=2$ for XYZ- and RSSI-based clustering approaches under the floor-based strategy, and $K=5$ for RSSI-based clustering under the building-based strategy. As $K$ increases beyond these values, the error distributions show an oscillatory increase, indicating unstable performance gains and a tendency toward over-segmentation. 

Overall, these results highlight that the hyperparameter $K$ plays a critical role in balancing model granularity and generalization: low values of $K$ tend to be sufficient in a more homogeneous radio environments, whereas more heterogeneous scenarios may benefit from a higher number of clusters.

\subsection{Computational cost analysis}
\label{subsec:comp_cost}
This subsection analyses the computational cost of the proposed method, focusing on the additional overhead introduced in both training and localisation phases by the clustering-based steps described in Section~\ref{subsec:cluster_estimation}.

\subsubsection{Training phase}
\begin{figure}[h]
    \centering
    \begin{subfigure}[b]{0.32\textwidth}
        \includegraphics[width=\linewidth, trim={0.8cm 0.8cm 1.95cm 1.5cm}, clip]{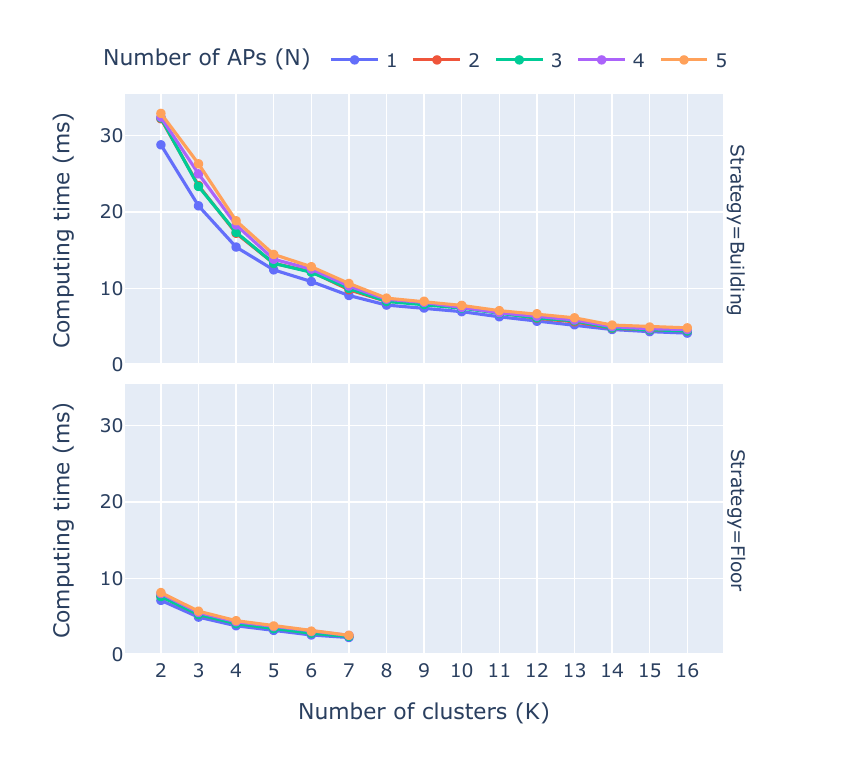}
        \caption{UJIIndoorLoc.}
        \label{subfig:uji_train_cost}
    \end{subfigure}
    \begin{subfigure}[b]{0.32\textwidth}
        \includegraphics[width=\linewidth, trim={0.8cm 0.8cm 1.95cm 1.5cm}, clip]{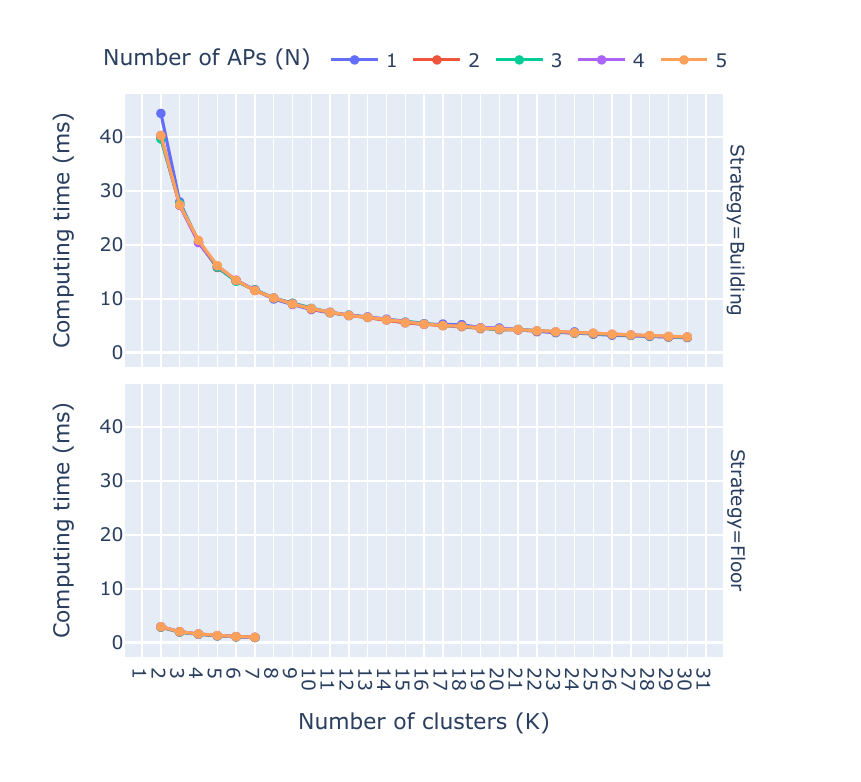}
        \caption{UTSIndoorLoc.}
        \label{subfig:uts_train_cost}
    \end{subfigure}
    \begin{subfigure}[b]{0.32\textwidth}
        \includegraphics[width=\linewidth, trim={{0.8cm 0.8cm 1.95cm 0.75cm}}, clip]{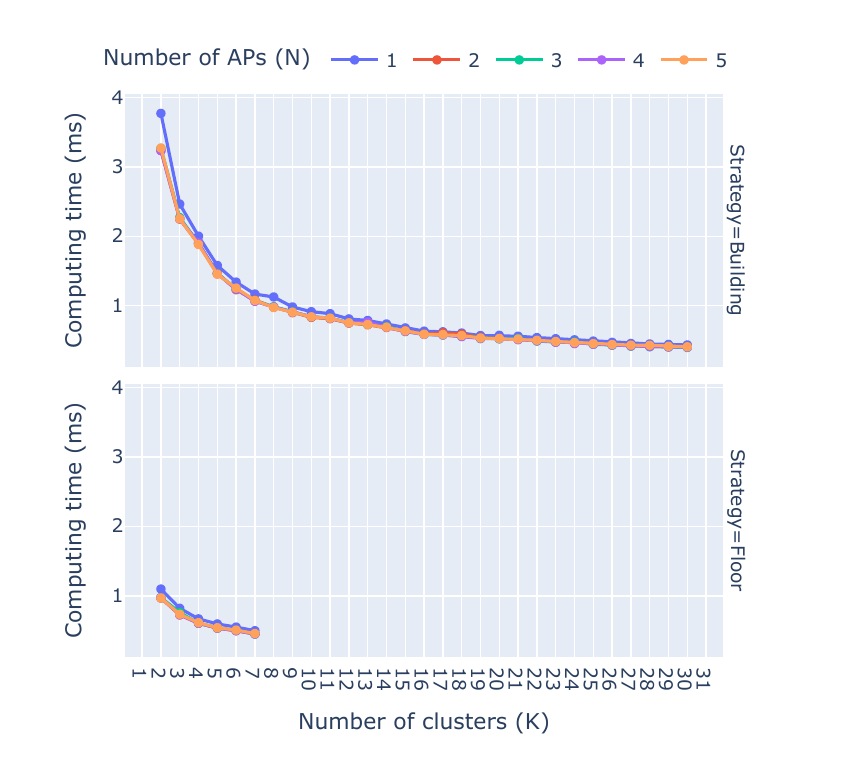}
        \caption{TUT.}
        \label{subfig:tut_train_cost}
    \end{subfigure}
    \caption{Mean computational cost in \unit{\milli\second} for constructing the representative AP sets during the training phase of the method. The upper subfigures correspond to the building-based strategy, whereas the lower subfigures correspond to the floor-based strategy.}
    \label{fig:train_cost}
\end{figure}

The training phase of the proposed method consists of two main components: cluster creation and model training. The cluster creation step involves two main operations. First, K-Means clustering is applied to partition the training data, which has a computational complexity of $\mathcal{O}(Kni)$, where $K$ is the number of clusters, $n$ is the number of training samples, and $i$ is the number of iterations required for convergence (limited to 300 iterations). The second operation consists of constructing the set of representative AP combinations per cluster, as defined in Equations~\ref{eq:cluster_representatives} and~\ref{eq:all_representatives}. Figure~\ref{fig:train_cost} shows the mean computational cost of building the representative AP sets for each cluster, depending on the $N$ and $K$ hyperparameters for each dataset. All datasets and both clustering strategies, a decreasing trend in computation time is observed as $K$ increases, while $N$ has little to no impact on the overall cost. The maximum mean overheads for constructing each cluster representative set are \SI{32.897}{\ms}, \SI{44.401}{\ms} and \SI{3.771}{\ms} when using two clusters, indicating that the cost is proportional to the amount of training samples.

Then, during model training, the impact of clustering depends on the selected ML model. For KNN-based models, clustering does not affect training time, as these models do not require explicit training beyond storing the reference fingerprints. For XGBoost and CNNLoc, clustering reduces the number of samples used to train each model, since training is performed independently within each cluster. As a result, the training time of these models is expected to decrease proportionally with the reduction in training data per cluster.

\subsubsection{Localisation phase}
\begin{figure}[h]
    \centering
    \begin{subfigure}[b]{0.32\textwidth}
        \includegraphics[width=\linewidth, trim={0.8cm 0.8cm 1.95cm 1.5cm}, clip]{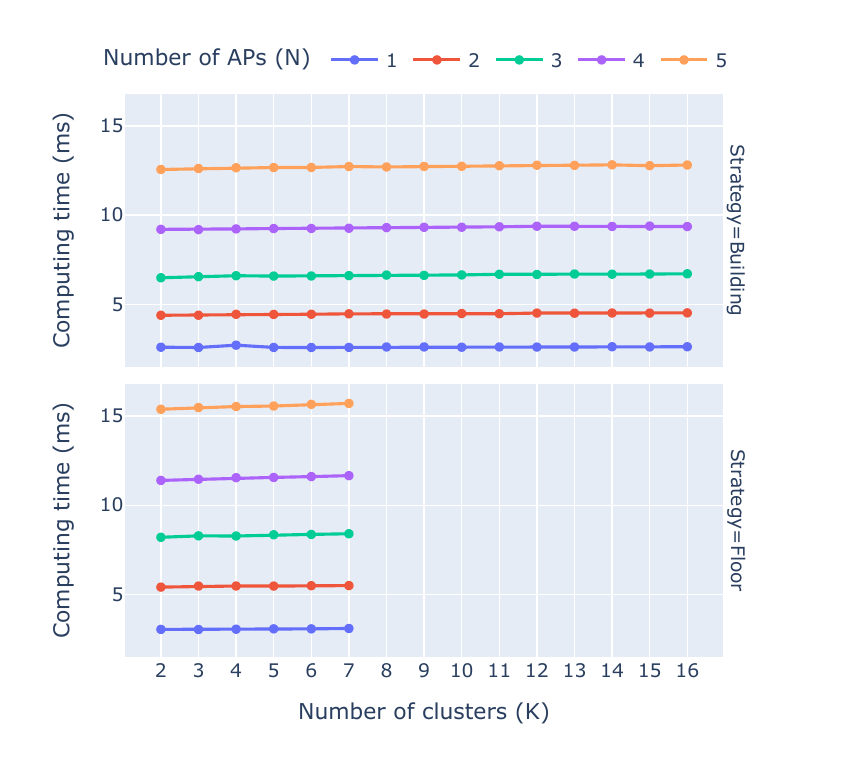}
        \caption{UJIIndoorLoc.}
        \label{subfig:uji_estimate_cost}
    \end{subfigure}
    \begin{subfigure}[b]{0.32\textwidth}
        \includegraphics[width=\linewidth, trim={0.8cm 0.8cm 1.95cm 1.5cm}, clip]{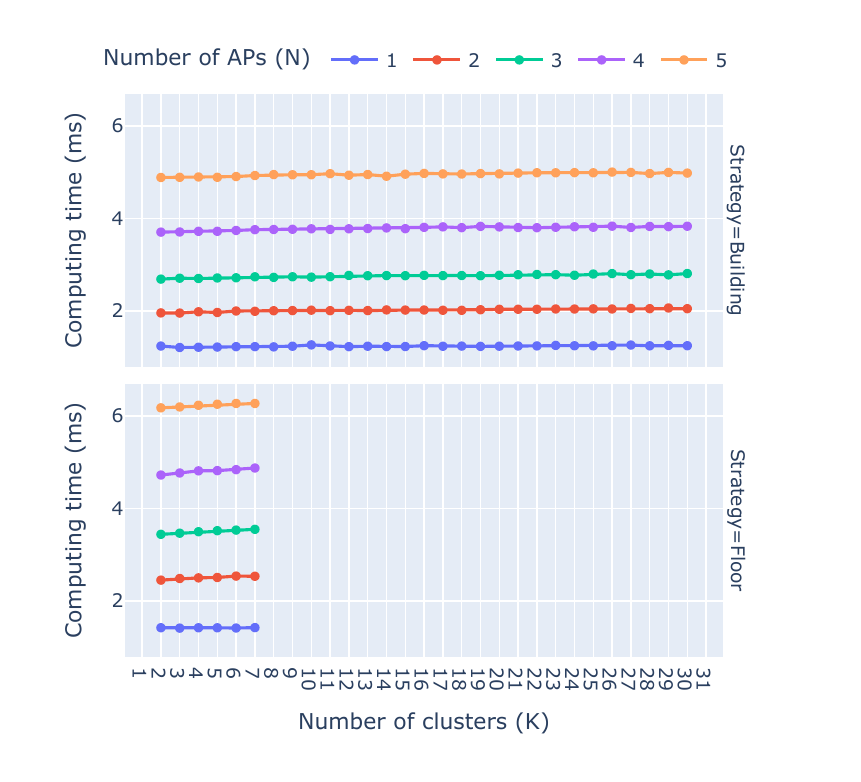}
        \caption{UTSIndoorLoc.}
        \label{subfig:uts_estimate_cost}
    \end{subfigure}
    \begin{subfigure}[b]{0.32\textwidth}
        \includegraphics[width=\linewidth, trim={{0.8cm 0.8cm 1.95cm 0.75cm}}, clip]{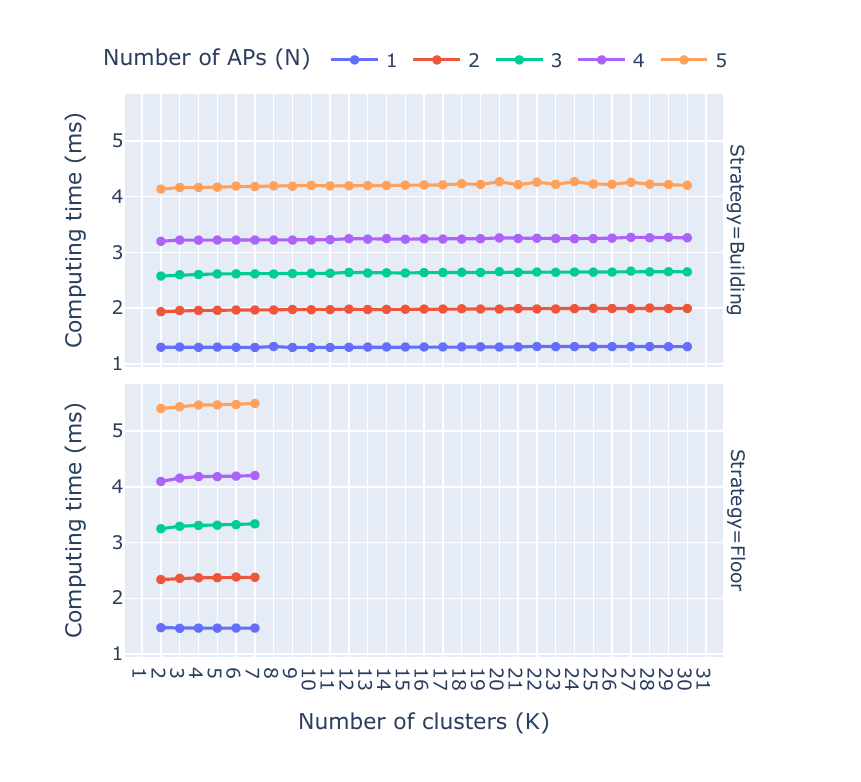}
        \caption{TUT.}
        \label{subfig:tut_estimate_cost}
    \end{subfigure}
    \caption{Mean computational cost in \unit{\milli\second} for estimating a sample to a cluster during the localisation phase of the method. The upper subfigures correspond to the building-based strategy, whereas the lower subfigures correspond to the floor-based strategy.}
    \label{fig:estimate_cost}
\end{figure}

The localisation phase includes two components: cluster estimation and model inference for each query sample. During localisation, the most suitable cluster for a given query is estimated using the $N$ selected APs, as defined in Section~\ref{subsec:cluster_estimation}. This step introduces an additional computational cost that depends on both $N$ and $K$, as depicted in Figure~\ref{fig:estimate_cost}. In this case, the main factor affecting the computational time overhead is $N$, where $N=1$ incurs in the lowest overhead, while $N=5$ results in the highest overhead. Regarding the hyperparameter $K$, it has little to no impact on the computational time overhead under the building-based strategy, whereas the floor-based strategy shows a slightly linear increase in cost as $K$ increases. Nevertheless, the overall overhead introduced by the cluster estimation step remains relatively low. For instance, the best performing models in each dataset from Section~\ref{subsec:pos_performance} would have had overhead of \SI{4.400}{\ms}\footnote{UJIIndoorLoc dataset, building-based strategy, RSSI-based approach, $N=2$, $K=2$.}, \SI{3.820}{\ms}\footnote{UTSIndoorLoc dataset, building-based strategy, RSSI-based approach, $N=4$, $K=17$.} and \SI{2.577}{\ms}\footnote{TUT dataset, building-based strategy, XYZ-based approach, $N=3$, $K=2$.} per sample.

Then, the impact of clustering on inference time depends on the learning models. For KNN-based approaches, clustering reduces the size of the search space, leading to faster distance computations. Therefore, the inference cost of KNN-based models is expected to decrease when clustering is applied. For XGBoost and CNNLoc, inference time is largely unaffected by clustering, as the computational cost remains similar regardless of the cluster size.

\subsection{Comparison with related work}\label{subsec:comparison}
Table~\ref{tab:comparison} reports a quantitative comparison between the proposed method and other state-of-the-art approaches that employ clustering techniques and KNN- and CNNLoc-based ML models across any of the three public datasets used in this work. To the best of our knowledge, no other works have used clustering techniques in conjunction with the XGBoost ML model. The comparison is conducted in terms of 2D positioning error ($e^{2D}$) and FDR.

\begin{table}[!h]
    \centering
    \caption{Comparison of 2D positioning errors (\unit{m}) and FDR (\unit{\percent}) against related works employing clustering-based approaches.}
    \input{tables/comparison}
    \label{tab:comparison}
\end{table}

For KNN-based approaches, the proposed method achieves competitive performance across all evaluated datasets. On UJIIndoorLoc, the best performing configuration (RSSI-based approach, building-based strategy) obtains an $e^{2D}$ of \SI{7.52}{\meter} with a FDR of \SI{93.53}{\percent}, which is comparable to other hierarchical and clustering-enhanced variants, such as the one proposed by \citet{liu2022hierarchical}, which presents the best results in this dataset. On UTSIndoorLoc, the proposed method (RSSI-based approach, building-based strategy) achieves an $e^{2D}$ of \SI{6.84}{\meter}, closely matching the best-performing (and only) method reported in the literature \cite{sebastian2025optimizing}. However, FDR values are not reported in that work. Regarding the TUT dataset, no clustering-based KNN methods report comparable results, and therefore, a direct comparison is not possible.

For CNNLoc-based approaches, the proposed method consistently outperforms existing methods across all evaluated datasets. On UJIIndoorLoc, the best performing configuration (RSSI-based approach, building-based strategy) achieves an $e^{2D}$ of \SI{8.13}{\meter}, while maintaining a competitive FDR of \SI{92.08}{\percent}, although slightly lower than those reported by other approaches~\cite{quezada2025modular,li2024hierarchical}. On the UTSIndoorLoc, the proposed method (XYZ-based approach, building-based strategy) achieves the lowest positioning error (\SI{7.43}{\meter}) and the highest FDR (\SI{94.07}{\percent}) among all compared methods. Similarly, on the TUT dataset, the proposed method (XYZ-based approach, building-based strategy) also attains the best reported results, with an $e^{2D}$ of \SI{9.72}{\meter} and a FDR of \SI{93.02}{\percent}, improving upon CAE-CNNLoc~\cite{kargar2024edge} and Modular CNNLoc~\cite{quezada2025modular}. This consistent improvement on UTSIndoorLoc and TUT datasets, compared with other works, indicates that the CNNLoc highly benefits more from spatial-based clustering than radio-based clustering.

Overall, the results demonstrate that the proposed method achieves state-of-the-art or near state-of-the-art performance across heterogeneous indoor environments and multiple ML models.

\section{Discussion}
\label{sec:6}
The proposed method demonstrates competitive performance across three heterogeneous datasets: UJIIndoorLoc, UTSIndoorLoc, and TUT. One of the main findings is the strong dependency of positioning error and FDR on the clustering approach (XYZ and RSSI) and the clustering strategy (building and floor).

The comparison of XYZ-based (i.e., spatial) and RSSI-based (i.e., radio) clustering approaches reveals that both exhibit complementary strengths, with the RSSI-based approach generally providing better positioning accuracy and stability across datasets (Tables~\ref{tab:uji},~\ref{tab:uts},~\ref{tab:tut}). For example, in UTSIndoorLoc, the RSSI-based approach achieves the best results in terms of positioning error and FDR for KNN-based models. However, the XYZ-based approach outperforms RSSI-based clustering in specific scenarios, particularly when using the CNNLoc model. This observation suggests that spatial-based clustering may provide richer spatial context that is more effectively exploited by complex learning models such as CNNLoc. This interpretation is consistent with the observed improvements obtained with XYZ-based approach when more advanced models are employed (e.g., XGBoost, CNNLoc), suggesting that encoding richer environmental structure, such as spatial distribution, can add value in more advanced learning architectures.

A consistent trend can be observed regarding the clustering strategies: the building-based strategy outperforms the floor-based strategy in most cases (Tables~\ref{tab:uji},~\ref{tab:uts},~\ref{tab:tut}). In both XYZ-based and RSSI-based approaches, the building-based strategy effectively reduces the 2D positioning errors and percentile errors, and improves FDR. In contrast, the floor-based strategy tends to yield poorer positioning accuracy and floor detection compared with the baseline method (i.e., no clustering). However, it is noticeable that the floor-based strategy improves the baseline method when no clustering is applied. This observation highlights the importance of choosing an appropriate strategy in IPS: while baseline methods can benefit from a floor-based strategy, clustering-based methods benefit more from a building-based strategy, providing a more favourable trade-off between positioning accuracy and floor detection performance than the floor-based strategy.

An important factor of the proposed method is the impact of its hyperparameters ($N$ and $K$) on overall performance. First, the number of APs (i.e., $N$) used for clustering significantly influences error variability, and increasing it from $1$ to $3$ generally provides the most substantial improvements in positioning accuracy (Figure~\ref{fig:n_boxplots}). Beyond this point, further increases in $N$ lead to diminishing returns, suggesting that using an excessive number of APs increases system complexity without corresponding gains in localisation accuracy. This finding reinforces the idea that a moderate number of APs ($\approx3$) is optimal for balancing complexity and performance. Second, the number of clusters (i.e., $K$) also plays a critical role: smaller values of $K$ tend to perform well for datasets such as UJIIndoorLoc and TUT, whereas a larger number of clusters yields better results for UTSIndoorLoc (Figure~\ref{fig:k_boxplots}). This variability in performance based on $K$ underscores the need for careful ``hyperparameter tuning'' depending on the environment and chosen clustering approach and strategy.

Overall, the empirical results indicate that the proposed method introduces a computational overhead during both cluster creation (training phase) and position estimation (localisation phase). During the localisation phase, where response time is critical, the overhead is directly determined by the hyperparameter $N$. Consequently, an appropriate trade-off between positioning accuracy and computational cost must be selected. Nevertheless, the positioning results demonstrate that the proposed method achieves improved localisation performance with a modest and controllable increase in computational cost. This additional cost can be partially compensated by the reduction in inference cost of models such as KNN-based approaches, due to the search space reduction enabled by clustering.


The results in Table~\ref{tab:comparison} show that the proposed method outperforms several state-of-the-art methods, specifically in CNNLoc-based models, where it obtains the best reported positioning errors across all datasets and the highest FDR values for the UTSIndoorLoc and TUT datasets. Regarding the results on the KNN-based models, the proposed method achieves comparable positioning accuracy but a lower FDR than the best performing state-of-the-art method~\cite{liu2022hierarchical}. However, the comparison for KNN-based approaches remains partial, as most existing works restrict their evaluation to a single dataset.

\subsection{Limitations}
Despite the reported results, the current study presents several limitations. First, the datasets used in the study, while widely adopted in the research community, may not fully represent the diversity of real-world indoor environments. Second, the trade-off between 2D positioning accuracy and FDR remains a critical challenge: the proposed method tends to improve positioning accuracy at the expense of FDR, which may limit its practical deployment in environments where accurate floor detection is crucial. This issue could potentially be mitigated through the integration of hybrid approaches, for example, by combining the floor detection capabilities of CNNLoc in baseline configurations with the 2D positioning accuracy of clustered KNN-based models. Finally, the comparison with related works is limited to studies applying clustering techniques and the same models and datasets considered in this work; therefore, other state-of-the-art approaches may not be included in the comparison.

\section{Conclusion}\label{sec:7}
This paper presented a novel approach for improving indoor positioning systems by integrating clustering techniques with ML models. The results demonstrated that both XYZ- and RSSI-based (i.e., spatial- and radio-based) clustering approaches offer significant improvements in positioning accuracy across multiple datasets, particularly when paired with ML models such as CNNLoc and XGBoost. The building-level strategy, when combined with clustering, consistently outperformed the floor-based strategy, providing a favourable balance between horizontal positioning accuracy and FDR. Despite achieving competitive performance in terms of 2D positioning error, the proposed method generally results in a reduced FDR compared with baseline approaches, highlighting the need for future improvements in balancing both metrics.

In summary, the proposed method showed promising potential for enhancing indoor localisation in heterogeneous environments. However, challenges remain, particularly in improving FDR while maintaining low positioning errors. Future research directions include extending the evaluation to additional datasets and learning models, exploring hybrid approaches that leverage the complementary strengths of individual models' benefits (e.g., the positioning accuracy of KNN-based models and the floor detection capabilities of CNNLoc), and developing adaptive strategies that account for environment-specific characteristics. By addressing these challenges, the proposed method can be made more robust and applicable to a wider range of real-world indoor environments.

\section*{Acknowledgement}
This work was funded by MCIN/AEI/10.13039/501100011033/ and ERDF/EU under Grant PID2022-141813OB-I00. It was also supported by a 2024 Leonardo Grant for Scientific Research and Cultural Creation from the BBVA Foundation.

\bibliographystyle{unsrtnat}
\footnotesize
\bibliography{references}  






\end{document}

%% file: tables/uji.tex
\begin{tabular}{llrrrrrrrrrrrrrrrrrrrrrrrrr}
\toprule
\multirow{2}{*}{\textbf{Model}} & \multirow{2}{*}{\textbf{Strat.}} &
\multicolumn{7}{c}{\textbf{Baseline}} &
\multicolumn{9}{c}{\textbf{XYZ Clusters}} &
\multicolumn{9}{c}{\textbf{RSSI Clusters}} \\ 
\cmidrule(lr){3-9} \cmidrule(lr){10-18} \cmidrule(lr){19-27}
& & $e^{2D}_{cf}$ & $e^{2D}_{50|cf}$ & $e^{2D}_{95|cf}$ & $e^{2D}$ & $e^{2D}_{50}$ & $e^{2D}_{95}$ & FDR &
$N$ & $K$ & $e^{2D}_{cf}$ & $e^{2D}_{50|cf}$ & $e^{2D}_{95|cf}$ &
$e^{2D}$ & $e^{2D}_{50}$ & $e^{2D}_{95}$ & FDR &
$N$ & $K$ & $e^{2D}_{cf}$ & $e^{2D}_{50|cf}$ & $e^{2D}_{95|cf}$ &
$e^{2D}$ & $e^{2D}_{50}$ & $e^{2D}_{95}$ & FDR \\
\midrule
KNN & Build. & 7.24 & \textbf{4.46} & 22.44 & 8.16 & \textbf{4.92} & 23.42 & \textbf{93.52} & 1 & 3 & 7.11 & 4.63 & 21.87 & 7.72 & \textbf{4.92} & 23.48 & 90.68 & 2 & 2 & \textbf{6.91} & 4.75 & \textbf{20.23} & \textbf{7.60} & 5.00 & \textbf{21.76} & 93.38 \\

WKNN & Build. & 7.22 & \textbf{4.47} & 22.45 & 8.14 & \textbf{4.84} & 23.36 & \textbf{93.52} & 1 & 3 & 7.09 & 4.63 & 21.79 & 7.70 & 4.94 & 23.52 & 90.68 & 2 & 2 & \textbf{6.88} & 4.74 & \textbf{20.24} & \textbf{7.56} & 4.98 & \textbf{21.66} & 93.47 \\

WKNN-T & Build. & 7.18 & \textbf{4.45} & 21.92 & 8.37 & \textbf{4.77} & 23.44 & 93.34 & 1 & 3 & 7.09 & 4.65 & 21.54 & 7.70 & 4.95 & 23.47 & 90.68 & 2 & 2 & \textbf{6.90} & 4.72 & \textbf{20.90} & \textbf{7.52} & 4.93 & \textbf{21.60} & \textbf{93.56} \\

XGBoost & Build. & 19.44 & 14.45 & 53.49 & 20.48 & 15.07 & 57.61 & 88.21 & 3 & 15 & 9.06 & \textbf{6.42} & 25.28 & 10.04 & \textbf{6.76} & 31.01 & \textbf{91.00} & 3 & 14 & \textbf{8.99} & 6.65 & \textbf{23.69} & \textbf{9.93} & 7.19 & \textbf{27.23} & 86.50 \\

CNNLoc & Build. & 9.37 & 7.66 & 21.62 & 10.08 & 7.99 & 24.01 & \textbf{92.89} & 1 & 3 & 8.39 & 6.39 & 21.16 & 8.82 & 6.65 & 22.71 & 90.59 & 3 & 4 & \textbf{7.64} & \textbf{6.11} & \textbf{20.01} & \textbf{8.13} & \textbf{6.38} & \textbf{21.35} & 92.08 \\ \midrule

KNN & Floor & \textbf{7.05} & \textbf{4.44} & \textbf{22.10} & 8.32 & \textbf{4.80} & \textbf{23.48} & \textbf{93.25} & 5 & 2 & 7.47 & 4.63 & 25.02 & 8.38 & 5.09 & 27.68 & 83.17 & 2 & 3 & 7.64 & 4.90 & 24.01 & \textbf{8.28} & 5.27 & 27.48 & 91.65 \\

WKNN & Floor & \textbf{7.05} & \textbf{4.44} & \textbf{21.99} & 8.30 & \textbf{4.76} & \textbf{23.45} & \textbf{93.25} & 5 & 2 & 7.46 & 4.66 & 25.07 & 8.37 & 5.06 & 27.35 & 83.17 & 2 & 3 & 7.60 & 4.84 & 24.01 & \textbf{8.25} & 5.11 & 27.37 & 91.65 \\

WKNN-T & Floor & \textbf{7.23} & \textbf{4.45} & \textbf{21.98} & 8.82 & \textbf{4.76} & \textbf{23.88} & \textbf{93.34} & 5 & 2 & 7.50 & 4.90 & 24.38 & 8.37 & 5.31 & 26.92 & 83.17 & 2 & 3 & 7.62 & 4.89 & 24.01 & \textbf{8.26} & 5.13 & 27.37 & 91.65 \\

XGBoost & Floor & 15.51 & 10.27 & 43.86 & 19.14 & 11.33 & 70.34 & 88.21 & 3 & 6 & \textbf{8.90} & \textbf{6.18} & \textbf{26.23} & \textbf{9.79} & \textbf{6.65} & 29.43 & \textbf{90.37} & 3 & 6 & 9.21 & 6.60 & 27.15 & 9.96 & 6.97 & \textbf{28.89} & 90.19 \\

CNNLoc & Floor & 23.29 & 15.36 & 54.98 & 23.95 & 15.99 & 57.84 & \textbf{95.05} & 3 & 6 & 8.48 & \textbf{6.09} & 24.23 & 9.27 & \textbf{6.29} & 26.74 & 90.37 & 2 & 2 & \textbf{8.29} & 6.32 & \textbf{22.28} & \textbf{8.91} & 6.65 & \textbf{24.13} & 91.38 \\
\bottomrule
\end{tabular}

%% file: tables/uts.tex
\begin{tabular}{llrrrrrrrrrrrrrrrrrrrrrrrrr}
\toprule
 \multirow{2}{*}{\textbf{Model}} & \multirow{2}{*}{\textbf{Strat.}} & \multicolumn{7}{c}{\textbf{Baseline}} & \multicolumn{9}{c}{\textbf{XYZ Clusters}} & \multicolumn{9}{c}{\textbf{RSSI Clusters}} \\ \cmidrule(lr){3-9} \cmidrule(lr){10-18} \cmidrule(lr){19-27}
 &  & $e^{2D}_{cf}$ & $e^{2D}_{50|cf}$ & $e^{2D}_{95|cf}$ & $e^{2D}$ & $e^{2D}_{50}$ & $e^{2D}_{95}$ & FDR & $N$ & $K$ & $e^{2D}_{cf}$ & $e^{2D}_{50|cf}$ & $e^{2D}_{95|cf}$ & $e^{2D}$ & $e^{2D}_{50}$ & $e^{2D}_{95}$ & FDR & $N$ & $K$ & $e^{2D}_{cf}$ & $e^{2D}_{50|cf}$ & $e^{2D}_{95|cf}$ & $e^{2D}$ & $e^{2D}_{50}$ & $e^{2D}_{95}$ & FDR \\
\midrule
KNN & Build.    & 7.15 & 6.25 & 15.36 & 8.20 & 6.79 & 19.94 & 85.31 & 4 & 3 & 7.10 & 6.32 & \textbf{14.96} & 8.08 & 6.99 & 18.82 & 84.54 & 4 & 17 & \textbf{6.53} & \textbf{5.43} & 15.15 & \textbf{7.11} & \textbf{5.87} & \textbf{18.17} & \textbf{87.11} \\
WKNN & Build.   & 7.16 & 6.23 & 15.54 & 8.20 & 6.81 & 19.95 & 85.57 & 4 & 3 & 7.11 & 6.37 & \textbf{14.93} & 8.08 & 6.90 & 18.88 & 84.79 & 4 & 17 & \textbf{6.54} & \textbf{5.53} & 15.33 & \textbf{7.13} & \textbf{5.90} & \textbf{18.07} & \textbf{87.11} \\
WKNN-T & Build. & 7.09 & 6.18 & 14.60 & 8.12 & 6.76 & 20.77 & 85.31 & 4 & 3 & 7.07 & 6.31 & 14.14 & 7.97 & 6.77 & 18.06 & 84.54 & 4 & 17 & \textbf{6.22} & \textbf{5.43} & \textbf{13.87} & \textbf{6.84} & \textbf{5.74} & \textbf{15.85} & \textbf{87.70} \\
XGBoost & Build.& 9.28 & 8.26 & 19.31 & 9.62 & 8.61 & 19.53 & \textbf{80.41} & 4 & 9 & \textbf{8.12} & 7.35 & \textbf{16.00} & \textbf{8.24} & \textbf{7.56} & \textbf{17.07} & 75.26 & 4 & 16 & 8.40 & \textbf{7.06} & 16.32 & 8.75 & \textbf{7.56} & 18.15 & 76.80 \\
CNNLoc & Build. & 7.56 & 6.92 & 15.83 & 7.95 & 7.05 & 16.76 & \textbf{95.88} & 4 & 2 & \textbf{7.03} & \textbf{6.29} & \textbf{15.41} & \textbf{7.43} & \textbf{6.51} & \textbf{16.26} & 94.07 & 4 & 11 & 7.49 & 6.48 & 17.46 & 7.67 & 6.59 & 18.37 & 92.78 \\ \midrule
KNN & Floor     & 7.09 & 6.22 & \textbf{15.68} & 8.14 & 6.93 & 19.60 & \textbf{86.34} & 5 & 2 & 7.70 & 6.63 & 18.21 & 8.58 & 6.95 & 21.00 & 79.12 & 4 & 6 & \textbf{6.96} & \textbf{5.63} & 17.56 & \textbf{7.80} & \textbf{6.42} & \textbf{19.26} & 77.98 \\
WKNN & Floor    & 7.11 & 6.21 & \textbf{15.74} & 8.15 & 6.98 & 19.50 & \textbf{86.34} & 5 & 2 & 7.70 & 6.76 & 18.35 & 8.59 & 7.01 & 21.10 & 79.12 & 4 & 6 & \textbf{6.96} & \textbf{5.67} & 17.64 & \textbf{7.81} & \textbf{6.42} & \textbf{19.32} & 77.98 \\
WKNN-T & Floor  & 6.93 & 6.19 & \textbf{14.78} & 8.00 & 6.77 & 19.15 & \textbf{85.57} & 5 & 2 & 7.43 & 6.28 & 17.36 & 8.36 & 6.84 & 20.13 & 79.37 & 4 & 6 & \textbf{6.64} & \textbf{5.58} & 16.96 & \textbf{7.62} & \textbf{6.19} & \textbf{18.54} & 78.69 \\
XGBoost & Floor & 9.06 & 7.38 & 22.12 & 10.36 & 8.26 & 27.62 & \textbf{80.67} & 5 & 6 & 8.05 & 7.02 & 19.09 & 8.90 & 7.65 & 21.02 & 77.06 & 4 & 6 & \textbf{7.95} & \textbf{6.93} & \textbf{18.18} & \textbf{8.57} & \textbf{7.49} & \textbf{20.73} & 77.84 \\
CNNLoc & Floor  & 12.55 & 10.05 & 35.49 & 12.91 & 10.53 & 35.49 & \textbf{94.33} & 4 & 6 & \textbf{8.12} & 7.25 & \textbf{17.92} & \textbf{8.82} & \textbf{7.45} & \textbf{18.98} & 78.35 & 4 & 4 & 8.52 & \textbf{6.82} & 18.74 & 9.33 & 7.80 & 22.27 & 77.58 \\
\bottomrule
\end{tabular}

%% file: tables/tut.tex
\begin{tabular}{llrrrrrrrrrrrrrrrrrrrrrrrrr}
\toprule
 \multirow{2}{*}{\textbf{Model}} & \multirow{2}{*}{\textbf{Strat.}} & \multicolumn{7}{c}{\textbf{Baseline}} & \multicolumn{9}{c}{\textbf{XYZ Clusters}} & \multicolumn{9}{c}{\textbf{RSSI Clusters}} \\ 
 \cmidrule(lr){3-9} \cmidrule(lr){10-18} \cmidrule(lr){19-27}
 &  & $e^{2D}_{cf}$ & $e^{2D}_{50|cf}$ & $e^{2D}_{95|cf}$ & $e^{2D}$ & $e^{2D}_{50}$ & $e^{2D}_{95}$ & FDR & $N$ & $K$ & $e^{2D}_{cf}$ & $e^{2D}_{50|cf}$ & $e^{2D}_{95|cf}$ & $e^{2D}$ & $e^{2D}_{50}$ & $e^{2D}_{95}$ & FDR & $N$ & $K$ & $e^{2D}_{cf}$ & $e^{2D}_{50|cf}$ & $e^{2D}_{95|cf}$ & $e^{2D}$ & $e^{2D}_{50}$ & $e^{2D}_{95}$ & FDR \\
\midrule
KNN & Build. & 9.65 & 6.46 & 28.85 & 10.36 & 6.82 & 31.73 & \textbf{89.60} & 3 & 2 & \textbf{8.68} & \textbf{6.29} & \textbf{23.74} & \textbf{9.55} & \textbf{6.66} & \textbf{27.04} & 89.11 & 1 & 5 & 9.81 & 6.80 & 28.62 & 10.67 & 7.29 & 31.94 & 87.46 \\
WKNN & Build. & 9.28 & 6.27 & 27.48 & 9.98 & 6.55 & 29.60 & \textbf{89.60} & 3 & 2 & \textbf{8.40} & \textbf{6.11} & \textbf{23.38} & \textbf{9.27} & \textbf{6.39} & \textbf{26.20} & 89.06 & 1 & 5 & 9.45 & 6.56 & 27.44 & 10.30 & 7.07 & 29.91 & 87.49 \\
WKNN-T & Build. & 9.31 & 6.30 & 27.50 & 9.98 & 6.55 & 29.60 & \textbf{90.26} & 3 & 2 & \textbf{8.49} & \textbf{6.14} & \textbf{23.52} & \textbf{9.27} & \textbf{6.39} & \textbf{26.29} & 89.85 & 1 & 5 & 9.49 & 6.60 & 27.52 & 10.29 & 7.07 & 29.91 & 88.20 \\
XGBoost & Build. & 10.39 & 8.19 & 26.13 & 10.82 & 8.50 & 27.52 & \textbf{90.03} & 5 & 3 & \textbf{8.99} & \textbf{6.62} & \textbf{24.04} & \textbf{9.30} & \textbf{6.71} & \textbf{25.17} & 86.33 & 1 & 2 & 10.21 & 8.20 & 25.82 & 10.88 & 8.64 & 28.57 & 86.57 \\
CNNLoc & Build. & 11.46 & 9.13 & 30.10 & 11.77 & 9.22 & 32.21 & 92.79 & 3 & 2 & \textbf{9.27} & \textbf{7.49} & \textbf{21.25} & \textbf{9.72} & \textbf{7.62} & \textbf{23.45} & 93.02 & 1 & 2 & 11.49 & 9.11 & 30.49 & 11.82 & 9.31 & 32.19 & \textbf{93.03} \\ \midrule
KNN & Floor & \textbf{9.25} & \textbf{6.06} & 28.57 & \textbf{10.61} & \textbf{6.37} & 32.40 & \textbf{89.50} & 5 & 2 & 9.84 & 6.86 & \textbf{26.97} & 11.36 & 7.75 & 32.41 & 82.25 & 3 & 2 & 10.31 & 7.05 & 29.58 & 11.97 & 7.93 & 36.69 & 82.89 \\
WKNN & Floor & \textbf{9.22} & \textbf{6.30} & 26.65 & \textbf{10.31} & \textbf{6.67} & 30.71 & \textbf{89.60} & 5 & 2 & 9.50 & 6.56 & \textbf{26.44} & 11.07 & 7.51 & 32.10 & 82.25 & 3 & 2 & 9.90 & 6.72 & 28.54 & 11.63 & 7.64 & 35.16 & 82.89 \\
WKNN-T & Floor & \textbf{9.23} & \textbf{6.32} & 27.02 & \textbf{10.03} & \textbf{6.66} & 29.61 & \textbf{90.26} & 5 & 2 & 9.50 & 6.56 & \textbf{26.44} & 11.07 & 7.51 & \textbf{31.90} & 82.24 &  3 & 2 & 9.89 & 6.72 & 28.54 & 11.61 & 7.64 & 35.06 & 82.89 \\
XGBoost & Floor & 9.45 & 7.00 & \textbf{25.14} & 10.64 & 7.46 & 31.27 & \textbf{89.83} & 5 & 2 & \textbf{9.24} & \textbf{6.78} & 25.09 & \textbf{10.64} & \textbf{7.38} & \textbf{31.11} & 82.25 & 3 & 2 & 9.49 & 6.97 & 25.85 & 11.06 & 7.70 & 30.81 & 82.82 \\
CNNLoc & Floor & 24.94 & 21.89 & 51.63 & 25.32 & 22.07 & 53.20 & \textbf{94.25} & 4 & 6 & \textbf{11.15} & \textbf{7.98} & \textbf{30.74} & \textbf{12.05} & \textbf{8.68} & \textbf{32.66} & 81.63 & 5 & 5 & 13.33 & 10.27 & 34.67 & 14.21 & 11.04 & 37.02 & 81.64 \\
\bottomrule
\end{tabular}

%% file: tables/comparison.tex
\begin{tabular}{llcccccc}
\toprule
 \multirow{2}{*}{\textbf{Model}} & \multirow{2}{*}{\textbf{Related Work}} & \multicolumn{2}{c}{\textbf{UJIIndoorLoc}} & \multicolumn{2}{c}{\textbf{UTSIndoorLoc}} & \multicolumn{2}{c}{\textbf{TUT}} \\ \cmidrule(lr){3-4} \cmidrule(lr){5-6} \cmidrule(lr){7-8}
 &  & $e^{2D}$ & FDR & $e^{2D}$ & FDR & $e^{2D}$ & FDR \\
\midrule
\multirow{7}{*}{KNN-based} & AFARLS \cite{gan2018hybrid} & 7.96 & 94.76 & --- & --- & --- & --- \\
                     & K-Means + WKNN \cite{siyang2021wknn} & 8.54 & 91.09 & --- & --- & --- & --- \\
                     & Hierarchical + WKNN \cite{liu2022hierarchical} & \textbf{7.51} & \textbf{95.23} & --- & --- & --- & --- \\ 
                     & TSVD + K-Means + IDW \cite{khoo2022enhanced} & 11.31 & --- & --- & --- & --- & --- \\ 
                     & K-Means + HNSW + WKNN \cite{sebastian2025optimizing} & --- & --- & \textbf{6.71} & --- & --- & --- \\
                     \cmidrule(lr){2-8}
                     & Our best & 7.52 & 93.53 & 6.84 & \textbf{87.70} & \textbf{9.27} & \textbf{89.85} \\
\midrule
\multirow{5}{*}{CNNLoc-based} & C-CNNLoc \cite{oh2021c} & 8.87 & --- & --- & --- & --- & --- \\
                        & CAE-CNNLoc \cite{kargar2024edge} & 9.52 & 90.50 & 7.70 & 92.00 & 10.24 & 88.90 \\
                        & Modular CNNLoc \cite{quezada2025modular} & 8.38 & \textbf{92.98} & 8.68 & 85.83 & 11.87 & 88.61 \\
                        & Linked-CNNLoc \cite{li2024hierarchical} & 8.71 & 92.80 & 7.87 & 93.00 & --- & --- \\
                        \cmidrule(lr){2-8}
                        & Our best & \textbf{8.13} & 92.08 & \textbf{7.43} & \textbf{94.07} & \textbf{9.72} & \textbf{93.02} \\
\bottomrule
\end{tabular}